\documentclass[10pt,twocolumn,letterpaper]{article}

\usepackage{cvpr}              

\definecolor{cvprblue}{rgb}{0.21,0.49,0.74}
\usepackage[pagebackref,breaklinks,colorlinks,allcolors=cvprblue]{hyperref}
\usepackage{booktabs}
\usepackage{graphicx}
\usepackage{multirow}
\newtheorem{proposition}{Proposition}
\newtheorem{theorem}{Theorem}
\def\paperID{34330} 
\def\confName{CVPR}
\def\confYear{2026}

\title{Eigen-Value: Efficient Domain-Robust Data Valuation via Eigenvalue-Based Approach
}

\author{Youngjun Choi\\
Yonsei University\\
Seoul, South Korea\\
{\tt\small choiyj9803@yonsei.ac.kr}
\and
Joonseong Kang\\
Yonsei University\\
Seoul, South Korea\\
{\tt\small doongsae@yonsei.ac.kr}
\and
Sungjun Lim\\
Yonsei University\\
Seoul, South Korea\\
{\tt\small lsj9862@yonsei.ac.kr}
\and
Kyungwoo Song\thanks{denotes corresponding author}\\
Yonsei University\\
Seoul, South Korea\\
{\tt\small kyungwoo.song@yonsei.ac.kr}
}

\begin{document}
\maketitle
\begin{abstract}

Data valuation is central to data-centric AI, enabling efficient training and objective data pricing by assigning a value to each data point. Most existing methods estimate the effect of removing data points based on in-distribution (ID) validation performance, often failing to generalize to out-of-distribution (OOD) settings where data follow different patterns. Although OOD-aware methods exist, they incur substantial computational costs.
To address this, we propose \emph{Eigen-Value} (EV), a plug-and-play data valuation framework for OOD robustness using only ID data. EV approximates domain discrepancy, defined as the gap between ID and OOD loss, via eigenvalue ratios of the ID covariance matrix, and estimates each data point’s marginal contribution using perturbation theory. It integrates seamlessly with ID loss–based methods without additional training.
Experiments show that EV improves OOD robustness and ranking stability across real-world datasets while remaining computationally efficient, making it practical for large-scale data valuation under domain shift.
\end{abstract}    
\section{Introduction}
\label{sec:intro}

\begin{figure*}[t]
  \centering
   \includegraphics[width=1\linewidth]{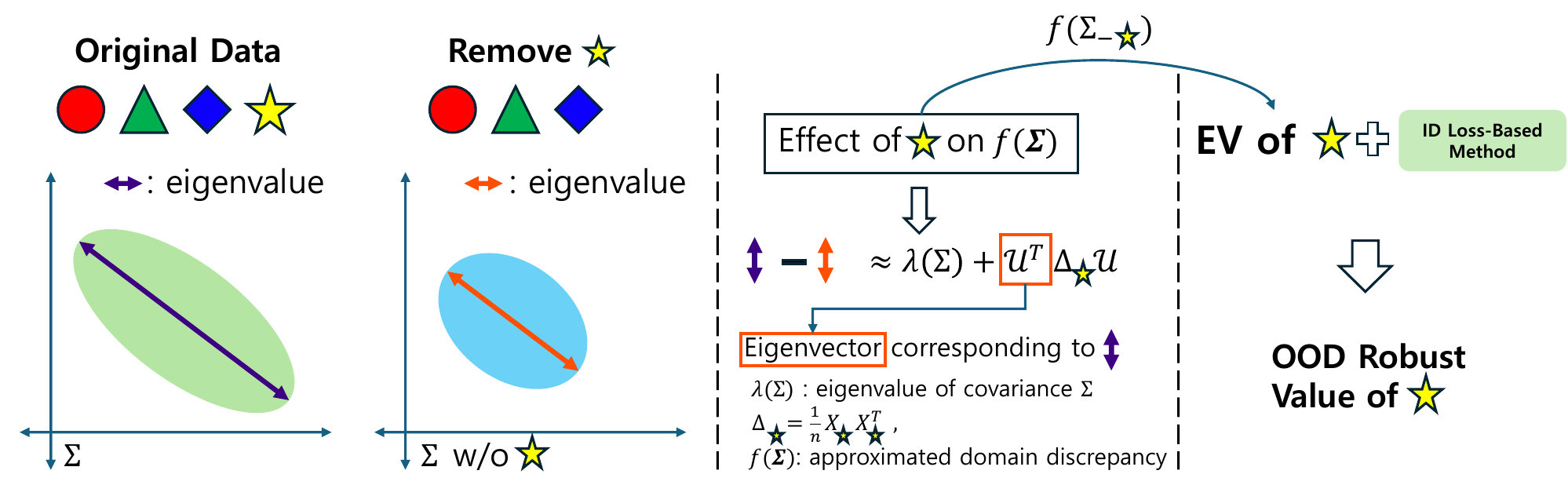}
   \caption{Overview of EV. Estimating the change in covariance eigenvalues induced by removing a single normalized embedding to quantify domain discrepancy, which is then integrated into ID loss-based data valuation for improved OOD robustness.
}
   \label{fig:overall_process}
\end{figure*}

Machine learning has achieved remarkable success in image recognition, autonomous driving, and conversational systems. However, domain shifts between in-distribution (ID) training data and out-of-distribution (OOD) deployment data can drastically degrade performance, posing risks in safety-critical settings. Most studies address this challenge through model-centric strategies such as distributionally robust optimization~\cite{hu2018does, staib2019distributionally, rahimian2022frameworks} or domain-invariant representation learning, often relying on specialized architectures and complex pipelines. A complementary data-centric paradigm, data valuation, identifies informative examples~\cite{sim2022data, tian2022private, agarwal2019marketplace}. Curating data rather than continually updating models reduces computation, accelerates learning, and enhances performance~\cite{xu2025quality}, while aligning with data marketplaces' pricing mechanisms.

Data valuation measures the change in model performance induced by including or excluding each training example. Most existing methods estimate this effect under the training distribution~\cite{shapley1953value, roth1988shapley, ghorbani2019data}. 
However, validation and OOD distributions often diverge, so values inferred from validation-based performance changes do not hold under OOD. This gap in distributional alignment makes OOD-robust data valuation necessary. However, existing shift-aware methods are computationally prohibitive~\cite{lin2024distributionally}, limiting their practicality in data marketplaces that require reliable metrics computed without OOD data~\cite{sim2022data, tian2022private}.

We address this gap with \emph{Eigen-Value} (EV), a data valuation framework for robustness under domain shift. EV connects domain discrepancy to the Hessian of the loss function, where in logistic regression the Hessian approximates data covariance~\cite{le1991eigenvalues, lin2007trust, hazan2014logistic}. Based on this, EV formulates domain discrepancy via eigenvalues of the covariance structure (Figure~\ref{fig:overall_process}). To reduce the cost of repeated eigendecomposition, it applies perturbation theory~\cite{kato2013perturbation} to approximate single-removal efficiently. EV extends ID loss–based valuation methods by incorporating a marginal value for eigenvalue shifts induced by domain discrepancy. Using ID data, EV estimates OOD robustness by quantifying how each sample perturbs the largest and smallest eigenvalues. Experiments show that augmenting baselines with EV improves OOD performance while maintaining efficiency.

In summary, our contributions are:
\begin{itemize}
\item We relate domain discrepancy to covariance eigenvalues, enabling data valuation without OOD samples.
\item We introduce EV, a scalable and easily integrable term that upgrades ID-based methods via perturbation theory.
\item We show on real-world datasets that EV improves OOD robustness, stability, and efficiency.
\end{itemize}
\section{Related Work}
\label{sec:related_work}

\subsection{Data Valuation}
Data valuation measures how each training example changes a model’s performance \cite{sim2022data}.
Its importance grows with large-scale datasets and retrieval augmented generation \cite{lewis2020retrieval}, where careful curation improves interpretability \cite{koh2017understanding}.
Data Shapley \cite{ghorbani2019data} estimates a point’s value by retraining across many subsets, which is expensive.
KNN Shapley \cite{jia2019efficient} and Data-OOB \cite{kwon2023data} reduce cost using local neighbors and out-of-bag scores, but face limits in scalability or accuracy.
LAVA \cite{just2023lava} removes the retraining bottleneck by measuring Wasserstein distances and gives fast estimates.
However, its effectiveness is limited to data sampled from the same source distribution as the model's training set, leading to degraded utility under domain shift.
Deviation~\cite{lin2024distributionally} formulates a worst-case distributional shift objective via the neural tangent kernel (NTK), enabling applicability beyond the training distribution. However, it requires \(n\) inversions of \(n \times n\) kernel matrix (with \(n\) samples), incurring huge computation that hampers practical deployment at scale. 
Even on reduced subsets, the method may exhibit instability, as its worst-case formulation~\cite{zhai2021doro} can amplify sensitivity to small changes, leading to fluctuations in value rankings and limiting suitability for data-centric AI at scale.
Our work departs from these existing methods and proposes an eigenvalue-based scheme that remains accurate under domain shift and is accelerated by perturbation theory.

\subsection{OOD Robustness}
OOD robustness seeks models that remain reliable when the test distribution deviates from that of the training and validation data
~\cite{ oh2024towards, wortsman2022robust}.
Distribution shifts, from small corruptions to large domain gaps, can reduce accuracy even for state-of-the-art networks, so training and evaluation must anticipate such changes.
Recent advances include ensembling and spectral criteria.
WiSE-FT \cite{wortsman2022robust} ensembles pre- and fine-tuned weights to improve both ID and OOD performance.
RankMe \cite{garrido2023rankme} and CaRoT \cite{oh2024towards} relate generalization to spectra of weight or feature matrices and guide optimization toward high rank or large minimum singular values.
These results suggest that spectral or weight space signals are useful proxies for OOD risk.
However, most of this literature on OOD robustness has been model-centric, and there has been a lack of an efficient data valuation method for assessing OOD loss. To enable efficient data pipeline management and provide objective pricing in data marketplaces, we propose an OOD-robust data valuation method. This data-centric approach allows us to evaluate OOD robustness from the perspective of data itself.

\subsection{Eigenvalue Methods}
Eigenvalue analysis has guided machine learning since classical PCA~\cite{hotelling1933analysis}.
Its view of covariance spectra also influenced kernel PCA, spectral clustering, and diffusion maps \cite{mackiewicz1993principal,scholkopf1997kernel,von2007tutorial,lafon2004diffusion}.
Recent works use spectral information to improve robustness.
CaRoT \cite{oh2024towards} encourages a larger smallest singular value of the ID input covariance matrix,
and Eigen-SAM \cite{luo2024explicit} discourages sharp directions to find flatter minima.
We extend this spectral line to data valuation.
By linking covariance eigenvalues to OOD generalization error and using perturbation theory to approximate each sample’s spectral influence, our method EV provides a scalable valuation approach aware of domain shift.

\section{Setting and Preliminaries}
\label{sec:setting_and_preliminaries}

\paragraph{Data Valuation.}
Let \(S=\{(x_i,y_i)\}_{i=1}^{N}\) be a dataset of \(N\) normalized
(embedding, label) pairs with \(x_i\in\mathcal{X}\subseteq\mathbb{R}^d\) and
\(y_i\in\mathcal{Y}\).
For any subset \(\mathcal{D}\subseteq S\) of size \(n\), a utility
function \(U:2^{n}\!\rightarrow\!\mathbb{R}\) maps the subset to a scalar
score.
We define the marginal data value
\(V:\mathcal{X}\times\mathcal{Y}\rightarrow\mathbb{R}\) of a point
\((x_k,y_k)\) as the drop in utility when that point is removed, i.e., $V(x_k,y_k)\;=\;U(\mathcal{D})
          \;-\;U\!\bigl(\mathcal{D}\setminus\{(x_k,y_k)\}\bigr)$.
Thus \(V(x_k,y_k)>0\) indicates that deleting the point hurts
performance, implying high importance.  
Existing methods share this core definition and differ only in how they approximate $U$ and compute $V$ efficiently.

\paragraph{Matching Marginal.}
Let $P_\text{ID}(x)$ denote the in-distribution (ID) used for training, and $P_\text{OOD}(x)$ an out-of-distribution (OOD) that differs only in the marginal $P(x)$ but shares the same conditional $P(y \mid x)$~\cite{shimodaira2000improving, moreno2012unifying}.
In other words, the input distribution shifts between ID and OOD, yet the relationship between input and label remains identical~\cite{sugiyama2007covariate}. 
To further analyze this OOD scenario, we consider the matching marginal condition. 
This condition implies that the covariance matrices of ID and OOD share the same diagonal entries, meaning that each feature has identical variance across domains. In contrast, the off-diagonal elements may differ~\cite{huang2006correcting}, reflecting possible shifts in cross-feature correlations.
Although normalization is not inherent to the definition, we apply feature-wise normalization to both ID and OOD data in our setting.
This step aligns the diagonal entries of their covariance matrices and makes the condition easier to satisfy.

\paragraph{Perturbation Theory.}
Perturbation theory offers a mathematical framework for approximating how a matrix’s eigenvalues and eigenvectors shift under a perturbation. For instance, if a matrix $A$ is modified by a small term $\epsilon$ such that $B=A+\epsilon$, one can express the eigenvalues and eigenvectors of $B$ as expansions in $\epsilon$~\cite{moro2003first,greenbaum2020first}. In the context of covariance matrices, this technique elucidates how adjustments to the underlying data affect eigenvalue structures \cite{sugiyama2020perturbation,mohammed2017perturbative}. Consequently, removing a single data point $k$ from a dataset can be viewed as applying a perturbation to the covariance matrix, thereby enabling an analysis of how individual samples influence eigenvalue shifts, as shown in Section~\ref{subsec:marginal_calculation_of_singular_value}. This perspective aids in evaluating dataset stability and identifying observations that exert outsized influence on the learned model.

\section{Efficient Domain Robust Data Valuation}
\label{sec:methodology}
We propose an eigenvalue-based framework for OOD-robust data valuation, focusing on the covariate-shift setting to ensure reliability under distribution shifts. First, we characterize domain discrepancy using eigenvalues of covariance matrix derived from normalized training dataset with zero mean $\mathcal{D}_\text{ID} = \{ z_i = (x_i, y_i) \}_{i=1}^{n}$, which is $\text{i.i.d.}$ sampled from the ID distribution to quantify shifts between ID and OOD data (Section~\ref{subsec:utility_function_based_on_ood_loss}). After that, we develop an efficient perturbation-based method to compute marginal valuations directly from eigenvalue terms, reducing computational overhead (Section~\ref{subsec:marginal_calculation_of_singular_value}). Finally, we integrate these components into a unified valuation framework, enabling scalable and robust data valuation without requiring explicit OOD samples (Section~\ref{subsec:final_expression}).

\subsection{Utility Function Based on OOD Loss}\label{subsec:utility_function_based_on_ood_loss}

In the Shapley value framework, the utility function \(U\), which forms the basis for data valuation, is typically defined as the model’s results on a validation set, with \(U(\mathcal{D})\) denoting validation performance on dataset \(\mathcal{D}\).
Similarly, we adopt \( \mathcal{L}_{\text{OOD}}(\theta) \) as our utility function, which is the loss function of a model with parameter \( \theta \)  trained on the ID set \(\mathcal{D}_\text{ID}\) and evaluated on OOD set \(\mathcal{D}_\text{OOD}\). The OOD loss \( \mathcal{L}_{\text{OOD}}(\theta) \) can be upper-bounded using the ID loss \( \mathcal{L}_{\text{ID}}(\theta) \) and a measure of domain discrepancy $\Gamma (\mathcal{D}_\text{OOD}, \mathcal{D}_\text{ID}) := \sup_\theta\left| \mathcal{L}_\text{OOD}(\theta) - \mathcal{L}_\text{ID}(\theta) \right|$. For notational simplicity, we denote $\mathcal{L}_\text{OOD}(\theta)$ and $\mathcal{L}_\text{ID}(\theta)$ as $\mathcal{L}_\text{OOD}$ and $\mathcal{L}_\text{ID}$.
\begin{equation}
  \mathcal{L}_\text{OOD} \le \mathcal{L}_\text{ID} + \Gamma(\mathcal{D}_\text{OOD}, \mathcal{D}_\text{ID})
  \label{eq:important}
\end{equation}

To quantify domain discrepancy, we first define it in terms of the distributional shift between ID and OOD data. A well-established approach is to approximate domain shift using a measure derived from the model’s sensitivity to input variations. Notably, when the loss function is formulated as Normalized Cross Entropy (NCE)~\cite{pmlr-v119-ma20c}, domain discrepancy can be related to the spectral properties of the Hessian matrices \( H_{\text{ID}} = \nabla^2_\theta \mathcal{L}_{\text{ID}}\) and \( H_{\text{OOD}} = \nabla^2_{\theta} \mathcal{L}_{\text{OOD}} \), corresponding to the ID and OOD distributions, respectively.

\begin{proposition}
\label{proposition_1}
We assume  \( \mathcal{L}_{\textnormal{ID}}\le\mathcal{L}_{\textnormal{OOD}} \) under the NCE loss function. Then, the domain discrepancy \( \Gamma(\mathcal{D}_{\textnormal{OOD}}, \mathcal{D}_{\textnormal{ID}}) \) is bounded as follows:

\begin{equation}
\label{eq:proposition_1}
\Gamma(\mathcal{D}_{\textnormal{OOD}}, \mathcal{D}_{\textnormal{ID}}) \leq \frac{\lambda_{\textnormal{max}}(H_{\textnormal{OOD}})}{\lambda_{\textnormal{min}}(H_{\textnormal{ID}})}
\end{equation}
\end{proposition}

where $\lambda_\textnormal{min}$ and $\lambda_\textnormal{max}$ stand for minimum eigenvalue and maximum eigenvalue. The derivation of Proposition~\ref{proposition_1} is provided in Appendix A.1. 
By representing domain discrepancy as the ratio of eigenvalues of the model, it becomes possible to leverage the characteristics of the logistic regression task to interpret information about the model's loss in terms of data under matching marginal assumptions.

In Proposition~\ref{proposition_1}, we represent domain discrepancy as a ratio of the eigenvalues $\frac{\lambda_{\textnormal{max}}(H_{\textnormal{OOD}})}{\lambda_{\textnormal{min}}(H_{\textnormal{ID}})}$. Furthermore, we express the ratio of a model’s eigenvalues in terms of the eigenvalues of the data. We can exploit the properties of logistic regression, where the Hessian of the loss corresponds to the data’s covariance matrix. 
As a result, we can formulate the notion of domain discrepancy not in terms of the model but rather using the eigenvalues of the data’s covariance matrix, allowing us to assess OOD robustness without relying on model-specific loss functions.

However, in practice, we usually lack information about OOD data. So we have no direct way of computing the eigenvalue for the OOD covariance matrix, which is used to approximate the numerator of the eigenvalue ratio in Eq.~\ref{eq:proposition_1}. In this context, the matching marginal assumption implies that the OOD data's covariance matrix $\Sigma_\text{OOD}$ can be modeled as the ID data's covariance matrix $\Sigma_\text{ID}$ with a perturbation. By taking the Frobenius norm of both sides of this relationship and then applying the triangle inequality, we can leverage standard properties of the Frobenius norm to upper-bound the eigenvalue of $\Sigma_\text{OOD}$ by that of $\Sigma_\text{ID}$.

\begin{theorem}
\label{theorem_1}
We assume $\Sigma_{\textnormal{OOD}} = \Sigma_{\textnormal{ID}} + E$, where $E \in \mathbb{R}^{d \times d}$ has zero diagonal entries and non-zero off-diagonal elements representing domain discrepancies.
Based on this assumption, we derive the following bound on $\mathcal{L}_{\textnormal{OOD}}$ in terms of $\lambda_{\max}$, $\lambda_{\min}$, and the dimensionality $d$ of $\Sigma_\textnormal{ID}$.

\begin{equation}
    \mathcal{L}_{\textnormal{OOD}} \le \mathcal{L}_{\textnormal{ID}} + 
    \frac{\lambda_{\max}(\Sigma_{\textnormal{ID}})\times\sqrt{d} + \sqrt{d^2-d}}{\lambda_{\min}(\Sigma_{\textnormal{ID}})}
    \label{eq:bound_of_Loss_OOD}
\end{equation}
\end{theorem}

Through this formulation, the bound on 
$\mathcal{L}_{\text{OOD}}$ of Eq.~\ref{eq:important} is established in Theorem~\ref{theorem_1}, with its derivation given in Appendix A.2.

\subsection{Marginal Calculation of Eigenvalue term}
\label{subsec:marginal_calculation_of_singular_value}
The change in the eigenvalue term from Eq.~\ref{eq:bound_of_Loss_OOD}, arising when the k-th data point $x_k$ is absent, serves as our target measure for data valuation. However, directly computing the eigenvalue for every single data point to capture this change becomes highly inefficient. To streamline the process, we first calculate $\Sigma_{-k}$, the covariance matrix without data point $x_k$, by perturbing the covariance matrix with the ID data subset $\Sigma_\textnormal{ID} = \frac{1}{n}\sum_i x_i x_i^\top$ with $\Delta_k= -\frac{1}{n}x_k\, x_k^\top$. Specifically, we can express it as:
\begin{equation}
\label{eq:cov_wo_k}
\Sigma_{-k} = \frac{1}{n-1}\sum_{i\neq k} x_i\,x_i^\top \approx \Sigma_\textnormal{ID} + \Delta_k
\end{equation}

Drawing on perturbation theory, we then approximate the eigenvalues of Eq.~\ref{eq:cov_wo_k} using the corresponding eigenvector $u$ as follows:
\begin{equation}
\label{eq:eign_of_pertubation_term}
\begin{split}
\lambda_{\text{max}}(\Sigma_{-k}) \approx \lambda_\text{max}(\Sigma_\textnormal{ID}+\Delta_k) &\approx \lambda_\text{max}(\Sigma_\textnormal{ID}) + u_\text{max}^\top\,\Delta_k\,u_\text{max}, \\
\lambda_{\text{min}}(\Sigma_{-k}) \approx \lambda_\text{min}(\Sigma_\textnormal{ID}+\Delta_k) &\approx \lambda_\text{min}(\Sigma_\textnormal{ID}) + u_\text{min}^\top\,\Delta_k\,u_\text{min}.
\end{split}
\end{equation}

\begin{figure}[t]
  \centering
   \includegraphics[width=1\linewidth]{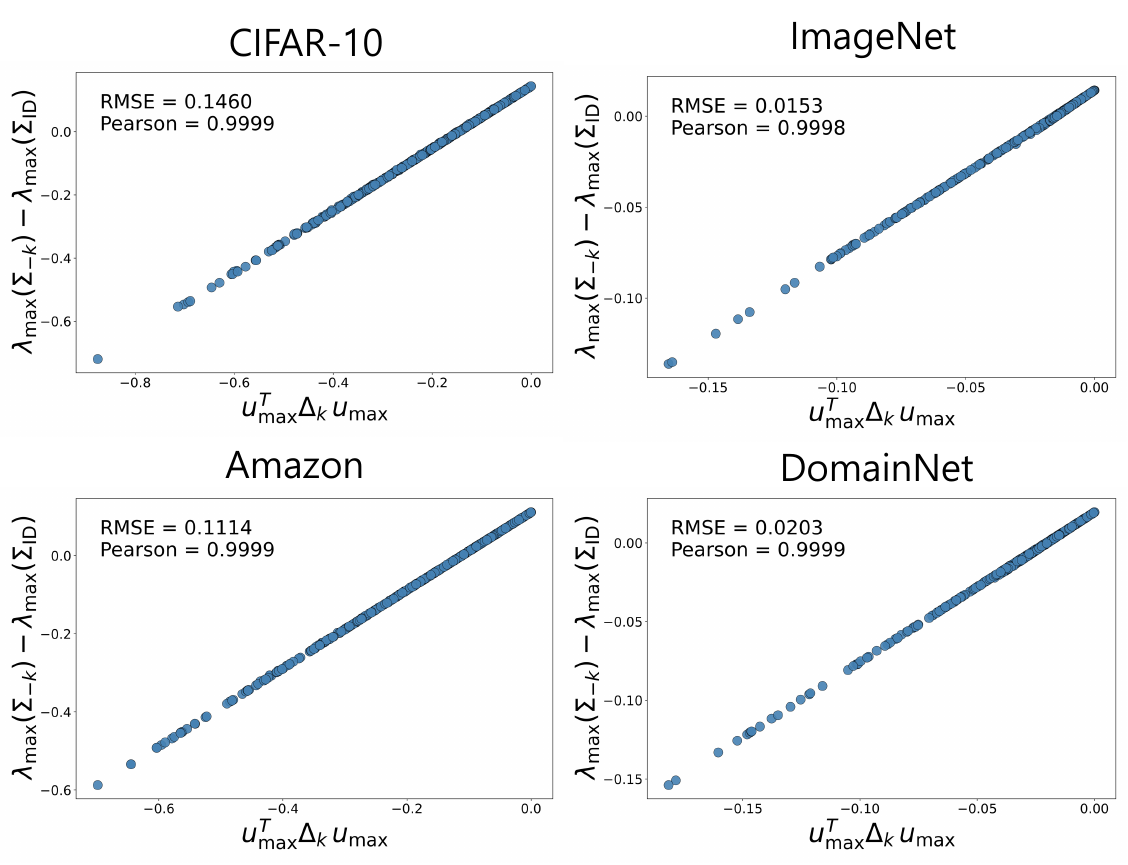}
   \caption{Relation between approximation values $u^\top_\text{max}\Delta_ku_\text{max}$ in Eq.~\ref{eq:eign_of_pertubation_term} and real values $\lambda_\text{max}(\Sigma_{-k}) - \lambda_\text{max}(\Sigma_\textnormal{ID})$ for CIFAR-10, ImageNet, Amazon Reviews - Books and DomainNet - Real embedding datasets. This demonstrates that eigenvalue differences can be accurately approximated using our method. The low RMSE and high Pearson correlation demonstrate the reliability of the approximation.}
   \label{fig:scatter_eigenvalue_approx}
\end{figure}

We denote $\delta_{\max}^{(k)}:= u_{\max}^\top \Delta_k u_{\max}$, $\delta_{\min}^{(k)}:= u_{\min}^\top \Delta_k u_{\min}$, capturing the sensitivity of the eigenvalues to the removal of data point \( x_k \) in Eq.~\ref{eq:eign_of_pertubation_term}.
To evaluate how well this approximation estimates the eigenvalue shift, we conducted an experiment using each 1k normalized embedding data point. Specifically, we examined the actual difference in eigenvalues with and without $x_k$ and assessed the linearity of our proposed approximation. 
Figure~\ref{fig:scatter_eigenvalue_approx} shows a clear linear trend between the true and estimated eigenvalue differences, supported by a low RMSE and a high Pearson correlation.
We observed that the difference in eigenvalues can be approximated using our proposed method.

Leveraging this insight, we approximate the difference in the eigenvalue term from Eq.~\ref{eq:bound_of_Loss_OOD} by directly substituting the approximation of the difference in eigenvalues.
More concretely, when considering the covariance matrix without data instance $k$, we replace its eigenvalue with  $\delta_{\max}^{(k)}, \delta_{\min}^{(k)} $. Then, by applying a Taylor expansion, we can use the resulting approximated terms, together with the eigenvalues of the covariance matrix, to approximate the marginal value of data point $x_k$ to the domain discrepancy.

\vspace{-1.2em}

\begin{theorem}
Let \( \lambda_{\max}(\Sigma_\textnormal{ID}) \) and \( \lambda_{\min}(\Sigma_\textnormal{ID}) \) be the maximum and minimum eigenvalues of the covariance matrix \( \Sigma_\textnormal{ID} \), and let \( \delta_{\max}^{(k)} \) and \( \delta_{\min}^{(k)} \) be the changes in these eigenvalues due to the perturbation caused by removing data point $x_k$. The marginal value of data point $x_k$ is then given by:
\begin{equation}
\begin{split}
f(\Sigma_{-k}) &- f(\Sigma_\textnormal{ID}) \approx \frac{\sqrt{d} \times \delta_{\max}^{(k)}}{\lambda_{\min}(\Sigma_\textnormal{ID})} \\
&- \frac{(\lambda_{\max}(\Sigma_\textnormal{ID}) \times \sqrt{d} + \sqrt{d^2 - d} \,) \times \delta_{\min}^{(k)}}{\lambda_{\min}(\Sigma_\textnormal{ID})^2} \nonumber
\end{split}
\end{equation}
where f($\Sigma_\textnormal{ID}$) is approximated domain discrepancy term in RHS of Eq.~\ref{eq:bound_of_Loss_OOD}, with its derivation given in Appendix A.3.
\end{theorem}

By leveraging this approximation, we can quantify the marginal value of each data point using perturbation theory without repeatedly performing costly eigendecompositions. This makes OOD-robust data valuation computationally efficient and feasible for practical deployment, achieving a complexity of $O(nd^2 + d^3)$, which is highly efficient in regimes where $n \gg d$.

\subsection{Eigen-Value: Plug and Play for ID Data Valuation Methodologies}\label{subsec:final_expression}
Since the calculation in Section~\ref{subsec:marginal_calculation_of_singular_value} pertains to the marginal value of domain discrepancy, it must be used in conjunction with the marginal value for ID loss. Leveraging this, we incorporated the eigenvalue-based term into existing data valuation methodologies for ID loss to perform marginal data value $V(x_k, y_k)$ for OOD loss,

\vspace{-1em}
\begin{equation}
\begin{split}
    V(x_k, y_k) &= \mathcal{L}_\text{ID}[S_{-k}]+f(\Sigma_{-k}) - f(\Sigma_\textnormal{ID}) \\
    &\approx \mathcal{L}_\text{ID}[S_{-k}]-\frac{\sqrt{d}\times\lambda_{\max}(\Sigma_\textnormal{ID}) + \sqrt{d^2 - d}}{\lambda_{\min}(\Sigma_\textnormal{ID})^2} \, \delta_{\min}^{(k)} \\
    &\quad + \frac{\sqrt{d}}{\lambda_{\min}(\Sigma_\textnormal{ID})} \, \delta_{\max}^{(k)}
\end{split}
\end{equation}
where $\mathcal{L}_\text{ID}[S_{-k}]$ is the marginal data value of other methods for ID loss.
\section{Experiments}
\label{sec:experiments}
We evaluate EV in three parts.
(1) \textbf{Cross-domain data removal and point addition.} We compute values on a source domain and measure performance on a different target domain to test whether the valuation is useful for selection.
(2) \textbf{Stability and efficiency.} To assess stability, we repeatedly alter a small subset of training samples and measure the variance in the resulting value rankings. We also compare computation time across methods. 
(3) \textbf{Qualitative analysis.} 
Top-ranked samples capture semantically stable class features and show wider dispersion in the embedding space, as seen in the qualitative top-3 examples. In contrast, low-ranked samples cluster redundantly and often lack such cues, explaining EV’s improved generalization under domain shift.
All valuations are computed using our implementation based on the OpenDataVal~\cite{jiang2023opendataval}.
Our code is available at 
\noindent\url{https://github.com/MLAI-Yonsei/Eigen-Value}.
\subsection{Experimental Settings}\label{subsec:experimental_settings}
\paragraph{Baselines.}\label{parap:baselines}
We compare our method, EV, with several baseline approaches for data valuation:  
(a) \textbf{Random}: Assigns data values randomly from a uniform distribution \( U(0,1) \).  
(b) \textbf{InfluenceFunction}~\cite{feldman2020neural}: Estimates the influence of an individual training example on the validation dataset by computing closely related sub-sampled influence.  
(c) \textbf{Deviation}: Compute data values using the distributionally robust generalization error (DRGE) based on NTK.  
(d) \textbf{LAVA}: A model-agnostic data valuation method that utilizes the class-wise Wasserstein distance.  
(e) \textbf{KNN Shapley}: Computes Shapley values using the K-Nearest Neighbors (KNN) approach.  
(f) \textbf{Data-OOB}: Employs the Out-of-Bag (OOB) technique to estimate data values.  
(g) \textbf{Eigen-Value}: Our proposed approach, EV, we conducted experiments applying other methods (LAVA, KNN Shapley, Data-OOB).
\begin{table*}[htbp]
\centering
\resizebox{0.9\textwidth}{!}{
\begin{tabular}{l|c|c c c c|c c c}
\toprule
\textbf{Acc(\%)} ($\downarrow$) 
& \multirow{2}{*}{\textbf{CIFAR-10 C}} 
& \multicolumn{4}{c|}{\textbf{VLCS}} 
& \multicolumn{3}{c}{\textbf{Amazon Reviews}} \\
\cmidrule(lr){1-1} \cmidrule(lr){3-6}\cmidrule(lr){7-9}
\textbf{Method} 
&  
& \textbf{Caltech101} & \textbf{LabelMe} & \textbf{SUN09} & \textbf{VOC2007} 
& \textbf{Books} & \textbf{Electronics} & \textbf{H and K} \\
\midrule
Random           & 47.01          &   95.90& 63.42 &   69.56  &  72.33&82.87&70.47&76.47  \\
InfluenceFunction  &  47.84     &  96.96& 62.36 &  70.81   &  67.97 &82.27&70.30&76.17 \\
Deviation             & 46.57     &  97.31& 62.10 &   68.12  &  73.48 &85.32&76.87&76.4 \\
LAVA             & 46.30          & 97.17 &  63.23 &  75.68  &  72.15  & 82.37 &70.55&76.50 \\
KNN Shapley              & 38.84         &  92.57& 59.04 &  55.85   &  52.39 &65.55&60.92&75.17  \\
Data-OOB              & 44.67          &  76.18& 62.70  &  62.94   &  56.72  &73.97&64.52&74.45\\
\midrule
EV + LAVA  & 43.96          &  93.78&62.48  & 49.60    & 70.17&81.62&70.35&75.95 \\
EV + KNN Shapley   & \textbf{38.68} &85.86& \textbf{58.14}& \textbf{49.08}  & \textbf{52.19}    &\textbf{55.85}&\textbf{47.92}&69.67 \\
EV + Data-OOB   & 43.65  &  \textbf{75.40}& 62.48    & 62.79 & 56.54   &56.15&48.12&\textbf{67.85}\\
\bottomrule
\end{tabular}
}
\caption{Data removal experiment. Train the model with 50\% of the data, which is the lowest data value in the ID set, and evaluate the performance on different domain data. \textbf{Lower is better.} The proposed method, which integrates EV with an existing approach, demonstrates strong performance. These results suggest that augmenting ID data valuation methods with EV provides a clearer guarantee of OOD performance compared to the Deviation approach. (*H and K stands for Amazon Reviews Home and Kitchen domain.)}
\label{tab:data_removal}
\end{table*}
\paragraph{Datasets.}\label{parap:dataset}
We conduct experiments on the following diverse datasets:  
(a) \textbf{CIFAR-10}~\cite{krizhevsky2009learning}: A widely used image classification dataset containing 60k images across 10 classes.  
(b) \textbf{CIFAR-10 C}~\cite{hendrycks2019benchmarking}: A variant of CIFAR-10 where common corruptions are applied, resulting in an image dataset with a distribution shift from the original data.  
(c) \textbf{VLCS}~\cite{fang2013unbiased}: A dataset comprising images from four distinct domains (VOC2007, LabelMe, Caltech101, SUN09), all sharing the same label space. 
(d) \textbf{Amazon Reviews}~\cite{hou2024bridging}: Amazon user product review data, which is organized into domains by product category. We convert the original 5-point ratings into three sentiment classes. In this work, we focus on the Books, Electronics, and Home and Kitchen domains.
(e) \textbf{ImageNet}~\cite{deng2009imagenet}: A large-scale dataset of labeled natural images spanning thousands of object categories, widely used for visual recognition research. In this work, we use several of its derived domains: ImageNet-V2 (V2)~\cite{recht2019imagenet}, ImageNet-Sketch (S)~\cite{wang2019learning}, ImageNet-R (R)~\cite{hendrycks2021many}, and ImageNet-A (A)~\cite{hendrycks2021natural}.
(f) \textbf{DomainNet}~\cite{peng2019moment}:  A benchmark dataset designed to evaluate cross-domain generalization, comprising images from six distinct domains covering the same set of object categories.

\paragraph{Setting.}\label{parap:setting}

Our method views the upper bound of OOD loss as a utility to pick ID samples that best improve cross-domain generalization. We computed EV scores from normalized ID embeddings, combined them with ID loss-based valuations. 
Based on the integrated scores, we curated a subset of the ID data to train a logistic regression model, which we optimized for 10 epochs using Adam with a learning rate of 0.001 and evaluated on the target domain.
For example, in VLCS, we designate SUN09 as the target domain. Data valuation and validation use the remaining three domains, and testing is performed solely on SUN09.
Throughout this process, we use embeddings extracted from either ResNet50~\cite{he2016deep} and ViT-B/16~\cite{dosovitskiy2020image} for image-based datasets, and RoBERTa-base~\cite{liu2019roberta} for text-based datasets such as Amazon Reviews.

\subsection{Cross Domain Experiment}\label{subsec:cross_domain_experiment}

\paragraph{Data Removal.}\label{parap:data_removal}

In the data removal experiment, we evaluate whether a valuation method can correctly identify low utility samples. From a pool of 2k ID data points, we randomly sample 1k for training and compute the value of each sample using each method. For every valuation baseline, we discard 50\% of samples assigned the highest values and train the model using only the remaining half. The model is then examined on OOD data. Since the retained training set is composed of data deemed less valuable, a larger accuracy drop indicates that the method was more effective at flagging uninformative samples. As shown in Table~\ref{tab:data_removal}, augmenting each baseline with EV consistently reduces the performance drop. Notably, EV + KNN Shapley achieves the best performance in all but two domains, where EV + Data-OOB outperforms all other approaches. These results highlight that EV provides a clearer guarantee of OOD robustness than existing alternatives such as Deviation.

To examine scalability beyond controlled small-scale settings and to test performance on more realistic and complex distributions, we further apply the same protocol to two large-scale benchmarks, ImageNet and DomainNet. Specifically, we subsample 30k images from ImageNet and 10k from DomainNet, compute per-sample values with all baselines in Table~\ref{tab:data_removal}, and repeat the data removal procedure. For scalability reasons, we omit LAVA and Deviation due to their prohibitive computational costs. On ImageNet, we evaluate across four domain shifts (V2, S, R, A), while on DomainNet we report results averaged across six shifts, with per-domain details provided in Appendix C.1. As summarized in Table~\ref{tab:data_removal_real}, the EV-augmented methods once again yield the lowest error in almost every target domain, confirming that our approach scales effectively and delivers superior robustness under substantial distribution shifts.

\begin{table}[!htbp]
\centering
\setlength{\tabcolsep}{0.8mm}
\begin{tabular}{l|cccc|c}
\toprule
\textbf{Acc(\%)} ($\downarrow$) 
& \multicolumn{4}{c|}{\textbf{ImageNet}}&\textbf{DomainNet} \\
\cmidrule(lr){1-1} \cmidrule(lr){2-5} \cmidrule(lr){6-6}
\textbf{Method}  & \textbf{V2} & \textbf{S} & \textbf{R} & \textbf{A} & \textbf{Avg.} \\
\midrule
Random               & 65.50 & 28.26 & 29.87 & 8.97&22.73 \\
InfluenceFunction    & 65.60 & 28.36 & 29.54 & 8.72&22.04 \\  
KNN Shapley          & 40.39 & 18.03 & 16.95 & 7.82&17.76 \\
Data-OOB             & 59.22 & 23.89 & 25.08 & 6.54 &11.76\\
\midrule
EV + KNN Shapley     & \textbf{40.34} & \textbf{17.97} & \textbf{16.91} & 7.81 &17.04\\
EV + Data-OOB        & 54.76 & 21.82 & 22.75 & \textbf{5.37} &\textbf{11.01}\\
\bottomrule
\end{tabular}
\caption{Data removal experiment. Train the model with 50\% of the data, which is the lowest data value in the ID set, and evaluate performance on different domain data. \textbf{Lower is better.} EV augmented variants consistently achieve the lowest error, which means EV achieves stronger OOD robustness than other methods. Because of their prohibitive time complexity on large, high-cardinality datasets, Deviation and LAVA are omitted.}
\label{tab:data_removal_real}
\vspace{-1.3em}
\end{table}

\paragraph{Point Addition.}\label{parap:data_addition}

\begin{figure*}[t]
  \centering
  \includegraphics[width=0.9\textwidth]{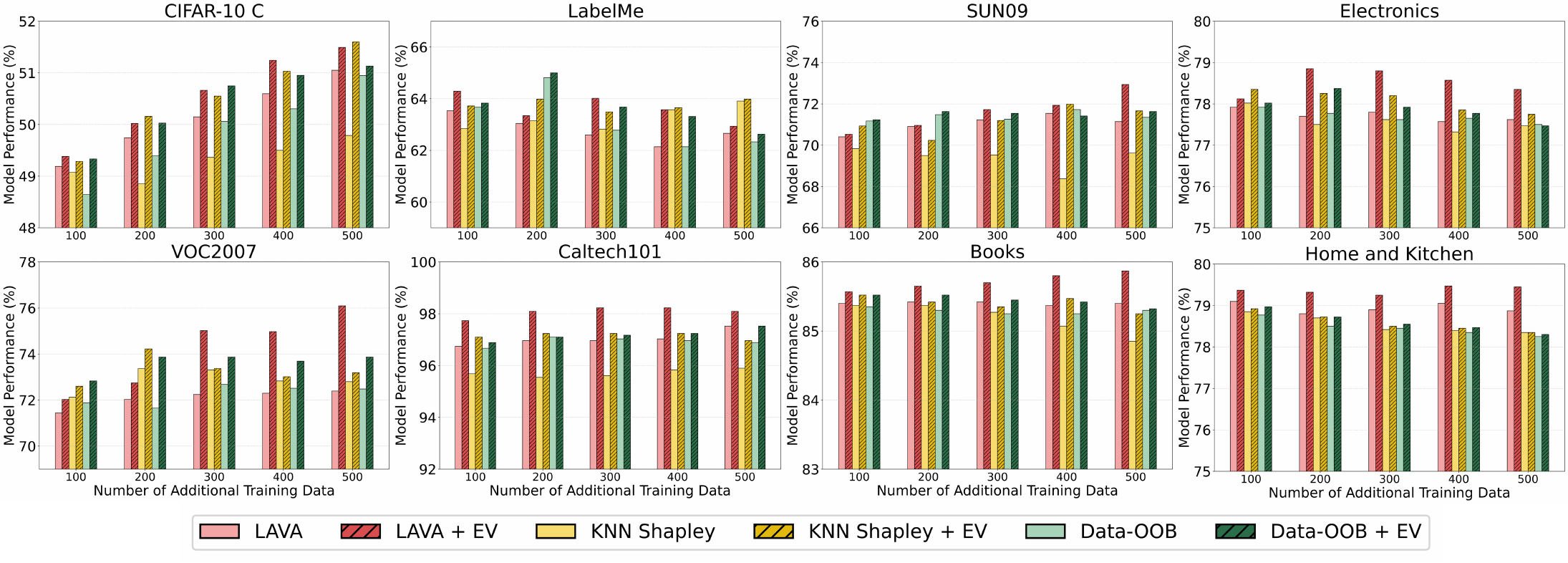}
  \caption{Performance comparison on OOD dataset, adding the highest data value of the remaining set. The hatched bars represent the performance of other methods when EV is applied. Results show that adding EV improves performance and enhances the robustness to OOD data. It highlights how selecting data based on our valuation approach can guide data inclusion in continual or online learning scenarios, where identifying the most beneficial data is crucial.}
  \label{fig:point_addition}
\end{figure*}

In the point addition experiment, we simulate a scenario where additional ID data are incrementally added to assess their effect on OOD generalization. We sample 2k ID data points, compute values with each method, and construct an initial 1k training set. The remaining 1k samples are then added in descending order of value, retraining the model after each step. This process is repeated on CIFAR-10 (tested on CIFAR-10 C), VLCS, and Amazon Reviews, covering both vision and text domains. Figure~\ref{fig:point_addition} shows performance after each addition, where solid bars denote baselines and hatched bars indicate their EV-augmented counterparts. Across domains, EV consistently achieves higher accuracy and stronger robustness to distribution shifts. These results confirm that EV effectively guides data selection toward informative and resilient samples, offering a principled criterion for continual or online learning, where prioritizing incoming data is critical.

\begin{figure}[t]
  \centering
   \includegraphics[width=0.9\linewidth]{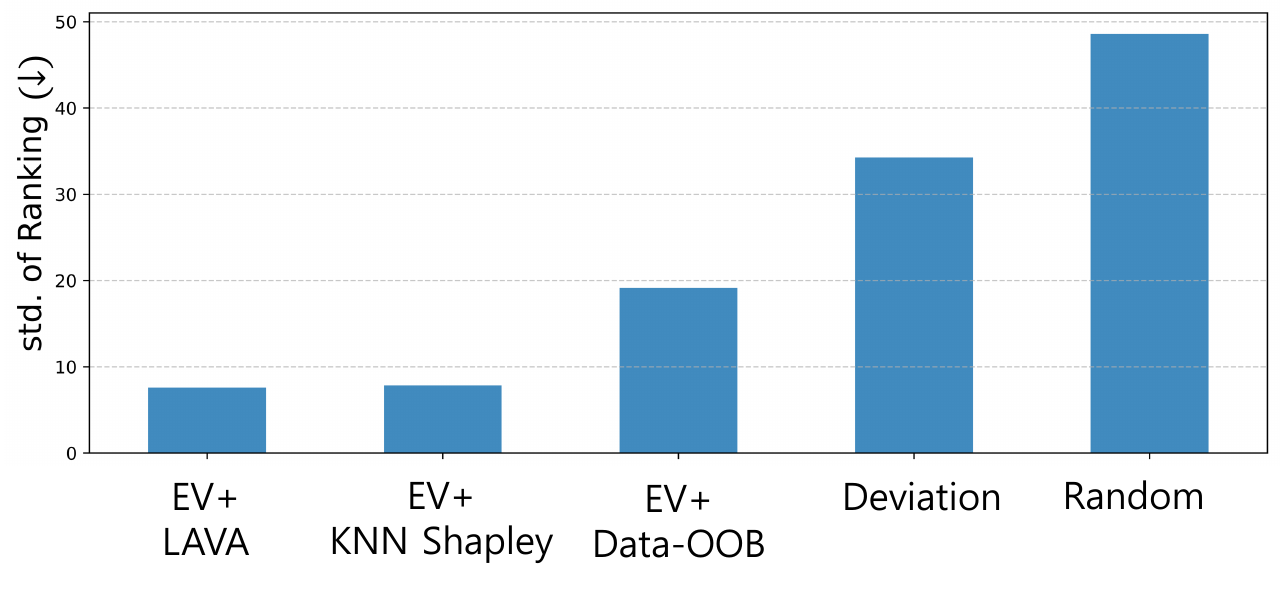}
   \caption{
Stability under training-set perturbations. We conduct a valuation on 200 CIFAR-10 samples five times, keeping 180 fixed and resampling 20. Deviation’s data-value ranking exhibits a standard deviation comparable to random selection, whereas EV yields stable, efficient rankings while retaining ID-based valuation and strong OOD performance.
   }
   \label{fig:Instability_bar_chart}
\end{figure}

\subsection{Stability and Efficiency of Eigen-Value}\label{subsec:stability_efficiency}

\paragraph{Instability Ranking.}\label{parap:instability_ranking}

A good data valuation method must be stable to small changes in the subset. Deviation estimates worst-case distribution error using NTK analysis by constructing a separate leave-one-out NTK matrix for each sample, which amplifies sensitivity to dataset composition and leads to unstable rankings. To evaluate stability, we fixed 180 out of 200 training samples and randomly replaced the remaining 20, repeating this procedure five times to compute the standard deviation of rankings. Ideally, if only a small portion of the dataset changes, the rankings of the fixed samples should vary within that range. However, as shown in Figure~\ref{fig:Instability_bar_chart}, Deviation exhibits much larger fluctuations, with the standard deviation of rankings approaching that of random selection. 
EV quantifies each sample’s contribution to domain discrepancy through covariance eigenvalues and produces consistent rankings. This stability makes EV more reliable for real-world data markets.

\begin{figure}[t]
  \centering
  \includegraphics[width=0.4\textwidth]{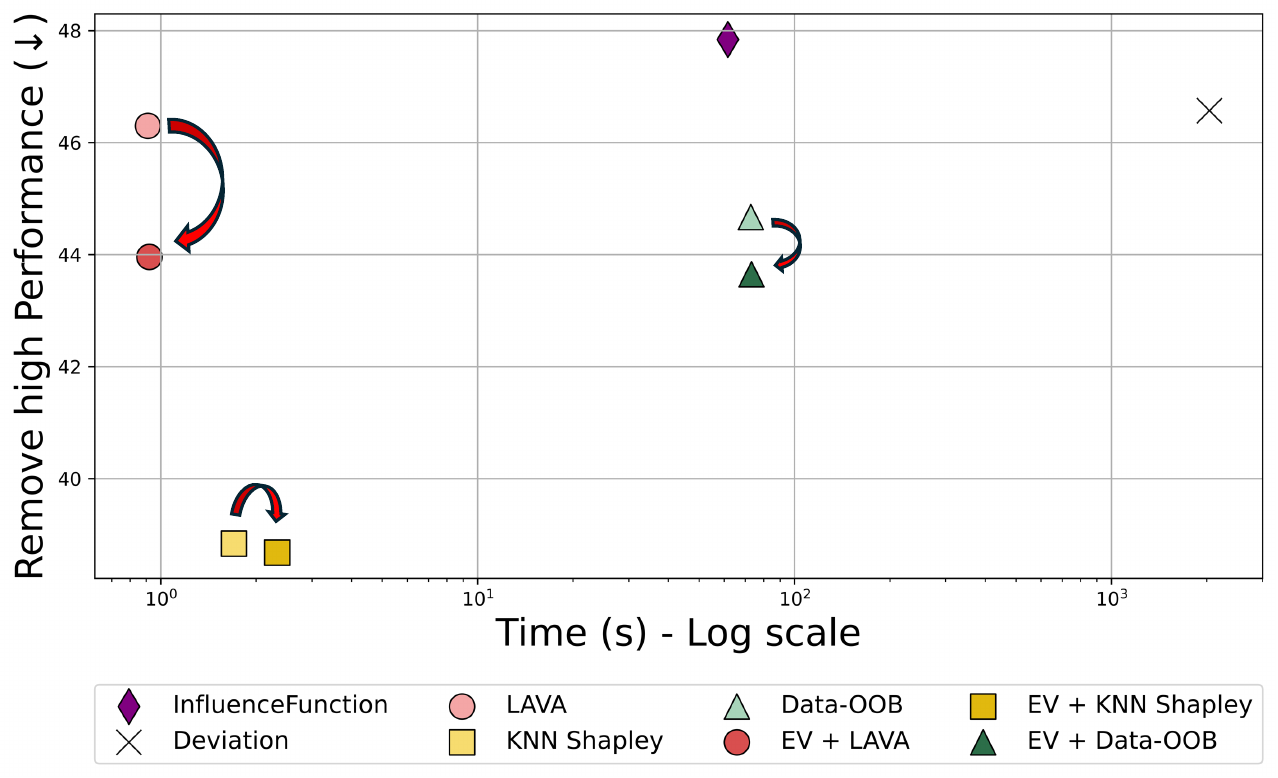}
  \caption{Time comparison on data valuation methods. Performance on CIFAR-10 C from Table 1, based on valuations and computation time using 2k samples of CIFAR-10. Despite its minimal overhead, EV outperforms Deviation in OOD robustness.
}
  \label{fig:time_comparison}
\end{figure}

\begin{figure*}[t]
  \centering
  \begin{subfigure}[t]{0.495\linewidth}
      \centering
      \includegraphics[width=\linewidth]{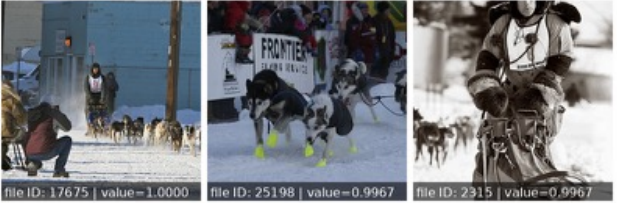}
      \caption{Top-3 samples in the dog sled class selected by Data-OOB after data valuation. Some images fail to capture invariant structures (e.g., dogs without sleds or unclear pulling).}
      \label{fig:qualitative_analysis_dataoob}
  \end{subfigure}
  \hfill
  \begin{subfigure}[t]{0.495\linewidth}
      \centering
      \includegraphics[width=\linewidth]{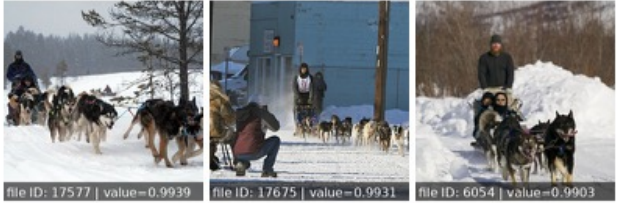}
      \caption{Top-3 samples in the dog sled class selected by EV + Data-OOB after data valuation. EV consistently prioritizes images where dogs are clearly pulling a sled.}
      \label{fig:qualitative_analysis_ev}
  \end{subfigure}
  
  \caption{Qualitative comparison of the Top-3 ranked images in the dog sled class from ImageNet, selected according to data values from (a) Data-OOB and (b) EV + Data-OOB. For each method, we first compute data values of images and then extract the three highest-value samples within the dog sled class.}
  \label{fig:qualitative_analysis}
\end{figure*}
\vspace{-1em}

\begin{figure}[t]
  \centering
   \includegraphics[width=1\linewidth]{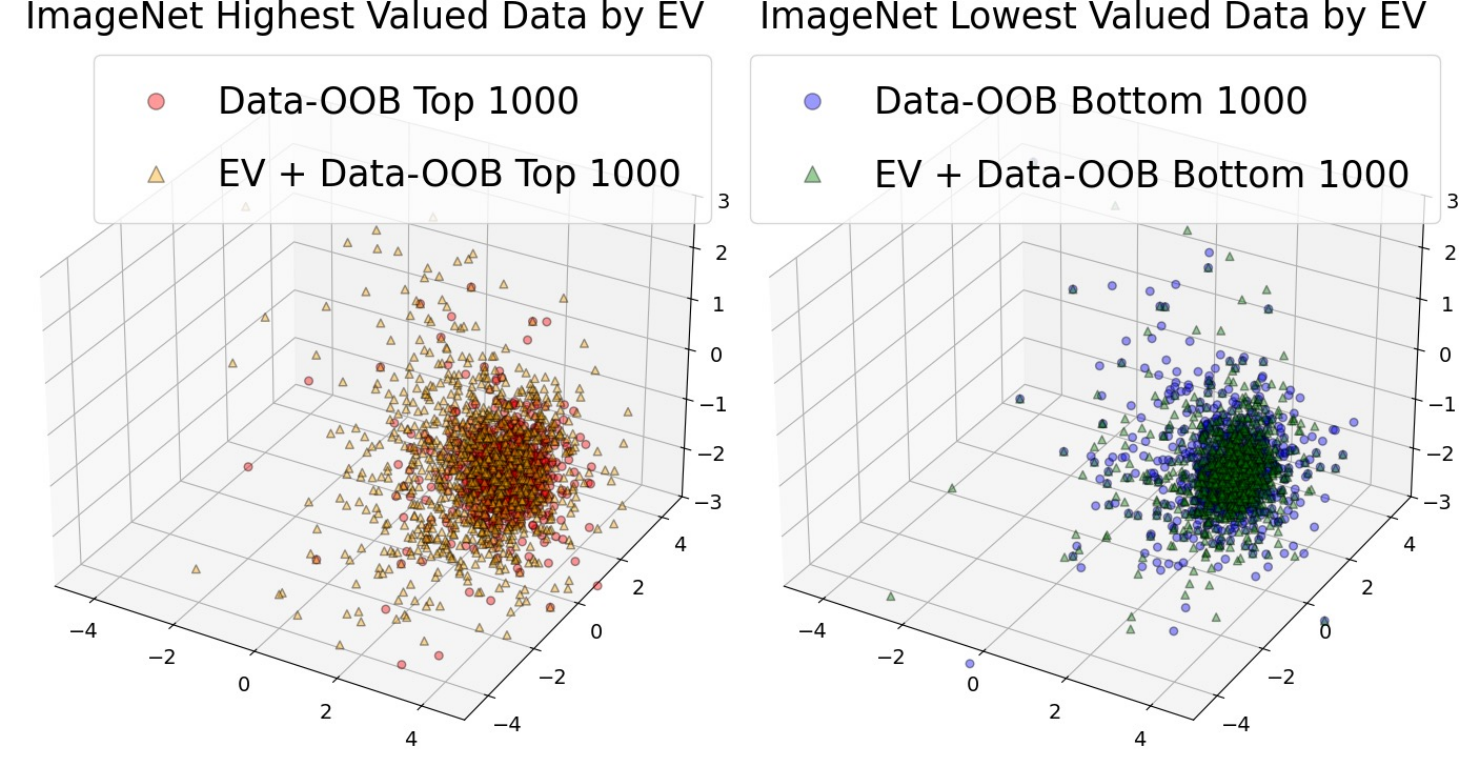}
   \caption{
   PCA projection of the top-1k and bottom-1k CIFAR-10 samples selected by Data-OOB with and without EV. Data-OOB alone produces an unexpected pattern where top samples have lower variance than bottom samples (0.38 vs. 0.60). With EV incorporated (EV + Data-OOB), the variance ordering becomes the opposite of what Data-OOB produces, with top samples showing larger dispersion than bottom samples (1.09 vs. 0.67). This helps the method identify broadly dispersed, high-variance samples that capture diverse features and improve OOD generalization.
   }
   \label{fig:pca_valued}
\end{figure}

\paragraph{Time Comparison.}\label{parap:time_comparison}
Another novelty of our approach is that it performs data valuation operations, yielding high performance within a short time. Although Deviation is a data valuation method that considers OOD data, it inverts an $n\times n$ matrix for each data point, resulting in a prohibitive computational complexity. This excessive computational burden, which grows cubically with dataset size, makes the method practically infeasible and limits its scalability in real-world applications. In contrast, our method adds only a small amount of additional computation compared to existing methods while demonstrating superior OOD performance. 
As shown in Figure~\ref{fig:time_comparison}, the methods augmented with EV achieve better OOD performance with minimal overhead of computing approximate eigenvalues for 2k samples, and take less than 1 second. In contrast, Deviation requires nearly 30 minutes and lacks OOD robustness.

\subsection{Qualitative Analysis}\label{subsec:qualitative_analysis}
\paragraph{Top-3 Sample Analysis with and without EV.}\label{parap:qualitative_analysis}
Previous quantitative experiments have demonstrated that EV improves domain robustness. To further investigate the reason, we conduct a qualitative analysis by examining which samples are assigned high data values. 
Specifically, we compare the three highest valued images selected by Data-OOB alone and by EV + Data-OOB.
Since domain robustness requires capturing invariant features essential for stability under domain shift, this comparison directly reveals how the two approaches differ in practice. We focus on the dog sled class in ImageNet, corresponding to the results in Table~\ref{tab:data_removal_real}. As shown in Figure~\ref{fig:qualitative_analysis}, Data-OOB often fails to capture invariant features; some images show only dogs, while others include a sled without clear pulling. In contrast, when EV is incorporated, the top-3 samples consistently highlight the defining invariant feature of the class, dogs visibly pulling a sled. This observation provides an intuitive explanation for why EV enhances OOD robustness, and similar patterns were observed across other classes as well.

\paragraph{Impact of EV.}\label{parap:impact_of_EV}

In Figure~\ref{fig:pca_valued}, we analyze the top-1k and bottom-1k ImageNet samples ranked by Data-OOB, with and without EV, and project them onto the top three principal components. For robust OOD performance, it is preferable to train on samples that are broadly distributed rather than narrowly clustered. However, Data-OOB without EV shows an inverted variance pattern in which top-ranked samples are more tightly clustered than bottom ones (Top: 0.38 vs. Bottom: 0.60), indicating that it may overlook diverse, informative examples. Adding EV restores the expected structure: top samples exhibit substantially higher dispersion than bottom samples (Top: 1.09 vs. Bottom: 0.67). This shows that EV prioritizes broadly distributed, high-variance samples, leading to a more representative training subset and stronger robustness under distribution shifts.
\section{Conclusion}
\label{sec:conclusion}
In this paper, we propose \emph{Eigen-Value} (EV), an efficient data valuation framework for OOD robustness. By approximating domain discrepancy via eigenvalues and perturbation theory, EV estimates each sample’s marginal contribution to OOD loss. Integrated with existing ID-based methods, it enables OOD-aware data selection without requiring additional OOD data. Extensive experiments on visual and textual datasets show that EV consistently enhances domain robustness while maintaining stability and low computational cost. Qualitative analyses further explain this improvement, showing that EV prioritizes diverse, invariant features critical under distribution shifts. By shifting the focus from model- to data-centric robustness, EV provides a scalable, theoretically grounded solution for real-world applications where efficient and reliable data valuation is essential.
\section*{Acknowledgment}
\label{sec:acknowledgment}
\raggedright
This work was supported by the National Research Foundation of Korea(NRF) grant funded by the Korea government(MSIT)(RS-2024-00457216).
{
    \small
    \bibliographystyle{ieeenat_fullname}
    \bibliography{main}

@InProceedings{pmlr-v119-ma20c,
  title = 	 {Normalized Loss Functions for Deep Learning with Noisy Labels},
  author =       {Ma, Xingjun and Huang, Hanxun and Wang, Yisen and Romano, Simone and Erfani, Sarah and Bailey, James},
  booktitle = 	 {Proceedings of the 37th International Conference on Machine Learning},
  pages = 	 {6543--6553},
  year = 	 {2020},
  editor = 	 {III, Hal Daumé and Singh, Aarti},
  volume = 	 {119},
  series = 	 {Proceedings of Machine Learning Research},
  month = 	 {13--18 Jul},
  publisher =    {PMLR},
  url = 	 {https://proceedings.mlr.press/v119/ma20c.html}
}

@article{tian2022private,
  title={Private data valuation and fair payment in data marketplaces},
  author={Tian, Zhihua and Liu, Jian and Li, Jingyu and Cao, Xinle and Jia, Ruoxi and Kong, Jun and Liu, Mengdi and Ren, Kui},
  journal={arXiv preprint arXiv:2210.08723},
  year={2022}
}

@inproceedings{agarwal2019marketplace,
  title={A marketplace for data: An algorithmic solution},
  author={Agarwal, Anish and Dahleh, Munther and Sarkar, Tuhin},
  booktitle={Proceedings of the 2019 ACM Conference on Economics and Computation},
  pages={701--726},
  year={2019}
}

@book{shapley1953value,
  title={A value for n-person games},
  author={Shapley, Lloyd S and others},
  year={1953},
  publisher={Princeton University Press Princeton}
}

@book{roth1988shapley,
  title={The Shapley value: essays in honor of Lloyd S. Shapley},
  author={Roth, Alvin E},
  year={1988},
  publisher={Cambridge University Press}
}

@inproceedings{ghorbani2019data,
  title={Data shapley: Equitable valuation of data for machine learning},
  author={Ghorbani, Amirata and Zou, James},
  booktitle={International conference on machine learning},
  pages={2242--2251},
  year={2019},
  organization={PMLR}
}

@inproceedings{
just2023lava,
title={{LAVA}: Data Valuation without Pre-Specified Learning Algorithms},
author={Hoang Anh Just and Feiyang Kang and Tianhao Wang and Yi Zeng and Myeongseob Ko and Ming Jin and Ruoxi Jia},
booktitle={The Eleventh International Conference on Learning Representations },
year={2023},
url={https://openreview.net/forum?id=JJuP86nBl4q}
}

@inproceedings{lin2024distributionally,
  title={Distributionally robust data valuation},
  author={Lin, Xiaoqiang and Xu, Xinyi and Wu, Zhaoxuan and Ng, See-Kiong and Low, Bryan Kian Hsiang},
  booktitle={Forty-first International Conference on Machine Learning},
  year={2024}
}

@inproceedings{hu2018does,
  title={Does distributionally robust supervised learning give robust classifiers?},
  author={Hu, Weihua and Niu, Gang and Sato, Issei and Sugiyama, Masashi},
  booktitle={International Conference on Machine Learning},
  pages={2029--2037},
  year={2018},
  organization={PMLR}
}

@article{staib2019distributionally,
  title={Distributionally robust optimization and generalization in kernel methods},
  author={Staib, Matthew and Jegelka, Stefanie},
  journal={Advances in Neural Information Processing Systems},
  volume={32},
  year={2019}
}

@article{rahimian2022frameworks,
  title={Frameworks and results in distributionally robust optimization},
  author={Rahimian, Hamed and Mehrotra, Sanjay},
  journal={Open Journal of Mathematical Optimization},
  volume={3},
  pages={1--85},
  year={2022}
}

@inproceedings{wu2022davinz,
  title={Davinz: Data valuation using deep neural networks at initialization},
  author={Wu, Zhaoxuan and Shu, Yao and Low, Bryan Kian Hsiang},
  booktitle={International Conference on Machine Learning},
  pages={24150--24176},
  year={2022},
  organization={PMLR}
}

@article{le1991eigenvalues,
  title={Eigenvalues of covariance matrices: Application to neural-network learning},
  author={Le Cun, Yann and Kanter, Ido and Solla, Sara A},
  journal={Physical review letters},
  volume={66},
  number={18},
  pages={2396},
  year={1991},
  publisher={APS}
}

@inproceedings{lin2007trust,
  title={Trust region newton methods for large-scale logistic regression},
  author={Lin, Chih-Jen and Weng, Ruby C and Keerthi, S Sathiya},
  booktitle={Proceedings of the 24th international conference on Machine learning},
  pages={561--568},
  year={2007}
}

@inproceedings{hazan2014logistic,
  title={Logistic regression: Tight bounds for stochastic and online optimization},
  author={Hazan, Elad and Koren, Tomer and Levy, Kfir Y},
  booktitle={Conference on Learning Theory},
  pages={197--209},
  year={2014},
  organization={PMLR}
}

@inproceedings{koh2017understanding,
  title={Understanding black-box predictions via influence functions},
  author={Koh, Pang Wei and Liang, Percy},
  booktitle={International conference on machine learning},
  pages={1885--1894},
  year={2017},
  organization={PMLR}
}

@inproceedings{yoon2020data,
  title={Data valuation using reinforcement learning},
  author={Yoon, Jinsung and Arik, Sercan and Pfister, Tomas},
  booktitle={International Conference on Machine Learning},
  pages={10842--10851},
  year={2020},
  organization={PMLR}
}

@inproceedings{sim2022data,
  title={Data Valuation in Machine Learning:" Ingredients", Strategies, and Open Challenges.},
  author={Sim, Rachael Hwee Ling and Xu, Xinyi and Low, Bryan Kian Hsiang},
  booktitle={IJCAI},
  pages={5607--5614},
  year={2022}
}

@article{shimodaira2000improving,
  title={Improving predictive inference under covariate shift by weighting the log-likelihood function},
  author={Shimodaira, Hidetoshi},
  journal={Journal of statistical planning and inference},
  volume={90},
  number={2},
  pages={227--244},
  year={2000},
  publisher={Elsevier}
}

@article{moreno2012unifying,
  title={A unifying view on dataset shift in classification},
  author={Moreno-Torres, Jose G and Raeder, Troy and Alaiz-Rodr{\'\i}guez, Roc{\'\i}o and Chawla, Nitesh V and Herrera, Francisco},
  journal={Pattern recognition},
  volume={45},
  number={1},
  pages={521--530},
  year={2012},
  publisher={Elsevier}
}

@article{sugiyama2007covariate,
  title={Covariate shift adaptation by importance weighted cross validation.},
  author={Sugiyama, Masashi and Krauledat, Matthias and M{\"u}ller, Klaus-Robert},
  journal={Journal of Machine Learning Research},
  volume={8},
  number={5},
  year={2007}
}

@article{huang2006correcting,
  title={Correcting sample selection bias by unlabeled data},
  author={Huang, Jiayuan and Gretton, Arthur and Borgwardt, Karsten and Sch{\"o}lkopf, Bernhard and Smola, Alex},
  journal={Advances in neural information processing systems},
  volume={19},
  year={2006}
}

@book{kato2013perturbation,
  title={Perturbation theory for linear operators},
  author={Kato, Tosio},
  volume={132},
  year={2013},
  publisher={Springer Science \& Business Media}
}

@article{moro2003first,
  title={First order eigenvalue perturbation theory and the Newton diagram},
  author={Moro, Julio and Dopico, Froil{\'a}n M},
  journal={Applied Mathematics and Scientific Computing},
  pages={143--175},
  year={2003},
  publisher={Springer}
}

@article{greenbaum2020first,
  title={First-order perturbation theory for eigenvalues and eigenvectors},
  author={Greenbaum, Anne and Li, Ren-cang and Overton, Michael L},
  journal={SIAM review},
  volume={62},
  number={2},
  pages={463--482},
  year={2020},
  publisher={SIAM}
}

@article{sugiyama2020perturbation,
  title={Perturbation theory approach to predict the covariance matrices of the galaxy power spectrum and bispectrum in redshift space},
  author={Sugiyama, Naonori S and Saito, Shun and Beutler, Florian and Seo, Hee-Jong},
  journal={Monthly Notices of the Royal Astronomical Society},
  volume={497},
  number={2},
  pages={1684--1711},
  year={2020},
  publisher={Oxford University Press}
}

@article{mohammed2017perturbative,
  title={Perturbative approach to covariance matrix of the matter power spectrum},
  author={Mohammed, Irshad and Seljak, Uro{\v{s}} and Vlah, Zvonimir},
  journal={Monthly Notices of the Royal Astronomical Society},
  volume={466},
  number={1},
  pages={780--797},
  year={2017},
  publisher={Oxford University Press}
}

@inproceedings{zhai2021doro,
  title={Doro: Distributional and outlier robust optimization},
  author={Zhai, Runtian and Dan, Chen and Kolter, Zico and Ravikumar, Pradeep},
  booktitle={International Conference on Machine Learning},
  pages={12345--12355},
  year={2021},
  organization={PMLR}
}

@article{oh2024towards,
  title={Towards calibrated robust fine-tuning of vision-language models},
  author={Oh, Changdae and Lim, Hyesu and Kim, Mijoo and Han, Dongyoon and Yun, Sangdoo and Choo, Jaegul and Hauptmann, Alexander and Cheng, Zhi-Qi and Song, Kyungwoo},
  journal={Advances in Neural Information Processing Systems},
  volume={37},
  pages={12677--12707},
  year={2024}
}

@inproceedings{garrido2023rankme,
  title={Rankme: Assessing the downstream performance of pretrained self-supervised representations by their rank},
  author={Garrido, Quentin and Balestriero, Randall and Najman, Laurent and Lecun, Yann},
  booktitle={International conference on machine learning},
  pages={10929--10974},
  year={2023},
  organization={PMLR}
}

@inproceedings{wortsman2022robust,
  title={Robust fine-tuning of zero-shot models},
  author={Wortsman, Mitchell and Ilharco, Gabriel and Kim, Jong Wook and Li, Mike and Kornblith, Simon and Roelofs, Rebecca and Lopes, Raphael Gontijo and Hajishirzi, Hannaneh and Farhadi, Ali and Namkoong, Hongseok and others},
  booktitle={Proceedings of the IEEE/CVF conference on computer vision and pattern recognition},
  pages={7959--7971},
  year={2022}
}

@book{lafon2004diffusion,
  title={Diffusion maps and geometric harmonics},
  author={Lafon, St{\'e}phane S},
  year={2004},
  publisher={Yale University}
}

@article{von2007tutorial,
  title={A tutorial on spectral clustering},
  author={Von Luxburg, Ulrike},
  journal={Statistics and computing},
  volume={17},
  pages={395--416},
  year={2007},
  publisher={Springer}
}

@inproceedings{scholkopf1997kernel,
  title={Kernel principal component analysis},
  author={Sch{\"o}lkopf, Bernhard and Smola, Alexander and M{\"u}ller, Klaus-Robert},
  booktitle={International conference on artificial neural networks},
  pages={583--588},
  year={1997},
  organization={Springer}
}

@article{mackiewicz1993principal,
  title={Principal components analysis (PCA)},
  author={Ma{\'c}kiewicz, Andrzej and Ratajczak, Waldemar},
  journal={Computers \& Geosciences},
  volume={19},
  number={3},
  pages={303--342},
  year={1993},
  publisher={Elsevier}
}

@article{lewis2020retrieval,
  title={Retrieval-augmented generation for knowledge-intensive nlp tasks},
  author={Lewis, Patrick and Perez, Ethan and Piktus, Aleksandra and Petroni, Fabio and Karpukhin, Vladimir and Goyal, Naman and K{\"u}ttler, Heinrich and Lewis, Mike and Yih, Wen-tau and Rockt{\"a}schel, Tim and others},
  journal={Advances in neural information processing systems},
  volume={33},
  pages={9459--9474},
  year={2020}
}

@article{feldman2020neural,
  title={What neural networks memorize and why: Discovering the long tail via influence estimation},
  author={Feldman, Vitaly and Zhang, Chiyuan},
  journal={Advances in Neural Information Processing Systems},
  volume={33},
  pages={2881--2891},
  year={2020}
}

@inproceedings{kwon2023data,
  title={Data-oob: Out-of-bag estimate as a simple and efficient data value},
  author={Kwon, Yongchan and Zou, James},
  booktitle={International conference on machine learning},
  pages={18135--18152},
  year={2023},
  organization={PMLR}
}

@article{jia2019efficient,
  title={Efficient task-specific data valuation for nearest neighbor algorithms},
  author={Jia, Ruoxi and Dao, David and Wang, Boxin and Hubis, Frances Ann and Gurel, Nezihe Merve and Li, Bo and Zhang, Ce and Spanos, Costas J and Song, Dawn},
  journal={arXiv preprint arXiv:1908.08619},
  year={2019}
}

@book{krizhevsky2009learning,
  title={Learning multiple layers of features from tiny images},
  author={Krizhevsky, Alex and Hinton, Geoffrey and others},
  year={2009},
  publisher={Toronto, ON, Canada}
}

@article{hendrycks2019benchmarking,
  title={Benchmarking neural network robustness to common corruptions and perturbations},
  author={Hendrycks, Dan and Dietterich, Thomas},
  journal={arXiv preprint arXiv:1903.12261},
  year={2019}
}

@inproceedings{fang2013unbiased,
  title={Unbiased metric learning: On the utilization of multiple datasets and web images for softening bias},
  author={Fang, Chen and Xu, Ye and Rockmore, Daniel N},
  booktitle={Proceedings of the IEEE international conference on computer vision},
  pages={1657--1664},
  year={2013}
}

@article{jiang2023opendataval,
  title={Opendataval: a unified benchmark for data valuation},
  author={Jiang, Kevin and Liang, Weixin and Zou, James Y and Kwon, Yongchan},
  journal={Advances in Neural Information Processing Systems},
  volume={36},
  pages={28624--28647},
  year={2023}
}

@inproceedings{he2016deep,
  title={Deep residual learning for image recognition},
  author={He, Kaiming and Zhang, Xiangyu and Ren, Shaoqing and Sun, Jian},
  booktitle={Proceedings of the IEEE conference on computer vision and pattern recognition},
  pages={770--778},
  year={2016}
}

@article{dosovitskiy2020image,
  title={An image is worth 16x16 words: Transformers for image recognition at scale},
  author={Dosovitskiy, Alexey and Beyer, Lucas and Kolesnikov, Alexander and Weissenborn, Dirk and Zhai, Xiaohua and Unterthiner, Thomas and Dehghani, Mostafa and Minderer, Matthias and Heigold, Georg and Gelly, Sylvain and others},
  journal={arXiv preprint arXiv:2010.11929},
  year={2020}
}

@article{hou2024bridging,
  title={Bridging Language and Items for Retrieval and Recommendation},
  author={Hou, Yupeng and Li, Jiacheng and He, Zhankui and Yan, An and Chen, Xiusi and McAuley, Julian},
  journal={arXiv preprint arXiv:2403.03952},
  year={2024}
}

@inproceedings{deng2009imagenet,
  title={Imagenet: A large-scale hierarchical image database},
  author={Deng, Jia and Dong, Wei and Socher, Richard and Li, Li-Jia and Li, Kai and Fei-Fei, Li},
  booktitle={2009 IEEE conference on computer vision and pattern recognition},
  pages={248--255},
  year={2009},
  organization={Ieee}
}

@inproceedings{peng2019moment,
  title={Moment matching for multi-source domain adaptation},
  author={Peng, Xingchao and Bai, Qinxun and Xia, Xide and Huang, Zijun and Saenko, Kate and Wang, Bo},
  booktitle={Proceedings of the IEEE/CVF international conference on computer vision},
  pages={1406--1415},
  year={2019}
}

@inproceedings{recht2019imagenet,
  title={Do imagenet classifiers generalize to imagenet?},
  author={Recht, Benjamin and Roelofs, Rebecca and Schmidt, Ludwig and Shankar, Vaishaal},
  booktitle={International conference on machine learning},
  pages={5389--5400},
  year={2019},
  organization={PMLR}
}

@inproceedings{hendrycks2021many,
  title={The many faces of robustness: A critical analysis of out-of-distribution generalization},
  author={Hendrycks, Dan and Basart, Steven and Mu, Norman and Kadavath, Saurav and Wang, Frank and Dorundo, Evan and Desai, Rahul and Zhu, Tyler and Parajuli, Samyak and Guo, Mike and others},
  booktitle={Proceedings of the IEEE/CVF international conference on computer vision},
  pages={8340--8349},
  year={2021}
}

@inproceedings{hendrycks2021natural,
  title={Natural adversarial examples},
  author={Hendrycks, Dan and Zhao, Kevin and Basart, Steven and Steinhardt, Jacob and Song, Dawn},
  booktitle={Proceedings of the IEEE/CVF conference on computer vision and pattern recognition},
  pages={15262--15271},
  year={2021}
}

@article{liu2019roberta,
  title={Roberta: A robustly optimized bert pretraining approach},
  author={Liu, Yinhan and Ott, Myle and Goyal, Naman and Du, Jingfei and Joshi, Mandar and Chen, Danqi and Levy, Omer and Lewis, Mike and Zettlemoyer, Luke and Stoyanov, Veselin},
  journal={arXiv preprint arXiv:1907.11692},
  year={2019}
}

@inproceedings{xu2025quality,
  title={Quality over quantity: Boosting data efficiency through ensembled multimodal data curation},
  author={Xu, Jinda and Song, Yuhao and Wang, Daming and Zhao, Weiwei and Chen, Minghua and Chen, Kangliang and Li, Qinya},
  booktitle={Proceedings of the AAAI Conference on Artificial Intelligence},
  volume={39},
  number={20},
  pages={21761--21769},
  year={2025}
}

@article{luo2024explicit,
  title={Explicit eigenvalue regularization improves sharpness-aware minimization},
  author={Luo, Haocheng and Truong, Tuan and Pham, Tung and Harandi, Mehrtash and Phung, Dinh and Le, Trung},
  journal={Advances in Neural Information Processing Systems},
  volume={37},
  pages={4424--4453},
  year={2024}
}

@article{hotelling1933analysis,
  title={Analysis of a complex of statistical variables into principal components.},
  author={Hotelling, Harold},
  journal={Journal of educational psychology},
  volume={24},
  number={6},
  pages={417},
  year={1933},
  publisher={Warwick \& York}
}

@article{wang2019learning,
  title={Learning robust global representations by penalizing local predictive power},
  author={Wang, Haohan and Ge, Songwei and Lipton, Zachary and Xing, Eric P},
  journal={Advances in neural information processing systems},
  volume={32},
  year={2019}
}

@article{samadi2018price,
  title={The price of fair pca: One extra dimension},
  author={Samadi, Samira and Tantipongpipat, Uthaipon and Morgenstern, Jamie H and Singh, Mohit and Vempala, Santosh},
  journal={Advances in neural information processing systems},
  volume={31},
  year={2018}
}

@article{das2024checkselect,
  title={CheckSelect: Online Checkpoint Selection for Flexible, Accurate, Robust, and Efficient Data Valuation},
  author={Das, Soumi and Sagarkar, Manasvi and Bhattacharya, Suparna and Bhattacharya, Sourangshu},
  journal={IEEE Transactions on Artificial Intelligence},
  volume={6},
  number={4},
  pages={968--978},
  year={2024},
  publisher={IEEE}
}

@inproceedings{wang2023data,
  title={Data banzhaf: A robust data valuation framework for machine learning},
  author={Wang, Jiachen T and Jia, Ruoxi},
  booktitle={International conference on artificial intelligence and statistics},
  pages={6388--6421},
  year={2023},
  organization={PMLR}
}
}

\clearpage
\setcounter{page}{1}
\maketitlesupplementary
\setcounter{section}{0}  
\renewcommand{\thesection}{\Alph{section}}

\renewcommand{\thesubsection}{\thesection.\arabic{subsection}}

\section{Theoretical Analysis}
This section provides the theoretical proof of EV, as detailed in Section 4 Efficient Domain Robust Data Valuation.

\subsection{Estimating Domain Discrepancy using Eigenvalue Shifts Induced by NCE}\label{subsec:a_1}

Normalized Cross-Entropy (NCE) is defined as
\begin{equation}
\text{NCE}(\theta) = \frac{-\sum_{k=1}^{K} q(y=k|x)\log p_\theta(k|x)}{-\sum_{j=1}^{K}\sum_{k=1}^{K} q(y=j|x)\log p_\theta(k|x)},
\tag{7}
\end{equation}
with \(0 \leq \text{NCE}(\theta) \leq 1\).

\noindent Since model parameter \(\theta\) is trained on in-distribution (ID) data, it is assumed that NCE on OOD data ($\text{NCE}_{\text{OOD}}$) is larger than NCE on ID data ($\text{NCE}_{\text{ID}}$) 
\[
0 < \text{NCE}_{\text{ID}}(\theta) \leq \text{NCE}_{\text{OOD}}(\theta)\leq 1.
\]

\medskip

\noindent Using the above relation, the domain discrepancy between OOD and ID can be defined as
\begin{equation}
\begin{split}
\Gamma(\mathcal{D}_{\text{OOD}}, \mathcal{D}_{\text{ID}})
&=\sup_{\theta} \Bigl( \text{NCE}_{\text{OOD}}(\theta) - \text{NCE}_{\text{ID}}(\theta) \Bigr) \\
&\leq \sup_{\theta} \frac{\text{NCE}_{\text{OOD}}(\theta)}{\text{NCE}_{\text{ID}}(\theta)}.
\end{split}
\tag{8}
\end{equation}

\medskip

\noindent Assuming an optimal model \(\theta_0\) for both domains, a Taylor expansion around \(\theta_0\) yields
\begin{equation}
\begin{split}
\text{NCE}(\theta) &\approx \text{NCE}(\theta_0) + (\theta - \theta_0)^\top \nabla_\theta \text{NCE}(\theta_0) \\
&\quad + \frac{1}{2} (\theta - \theta_0)^\top \nabla^2_\theta \text{NCE}(\theta_0)(\theta - \theta_0).
\end{split}
\tag{9}
\end{equation}
Since \(\text{NCE}(\theta_0) \approx 0\) and \(\nabla_\theta \text{NCE}(\theta_0) \approx 0\), it follows that
\[
\text{NCE}(\theta) \approx \frac{1}{2} (\theta - \theta_0)^\top H (\theta - \theta_0),
\]
where \(H := \nabla^2_\theta \text{NCE}(\theta_0)\).

\medskip

\noindent Thus, the ratio can be approximated by the Hessian of each distribution ($H_\text{OOD}, H_\text{ID}$)
\begin{equation}
\sup_{\theta} \frac{\text{NCE}_{\text{OOD}}(\theta)}{\text{NCE}_{\text{ID}}(\theta)}
\approx \sup_{\theta} \frac{\frac{1}{2} (\theta - \theta_0)^\top H_{\text{OOD}} (\theta - \theta_0)}
{\frac{1}{2} (\theta - \theta_0)^\top H_{\text{ID}} (\theta - \theta_0)}.
\tag{10}
\end{equation}
Using the Rayleigh quotient property, for any nonzero vector $v \in \mathbb{R}^d$
\[
\lambda_{\min}(H) \leq \frac{v^\top H\, v}{v^\top v} \leq \lambda_{\max}(H).
\]
Then, under the assumption that the Hessian is positive semi-definite, we approximate the ratio of NCE between distributions using the ratio of their maximum ($\lambda_{\max}$) and minimum ($\lambda_{\min}$) eigenvalues.

\begin{equation}
\frac{\text{NCE}_{\text{OOD}}(\theta)}{\text{NCE}_{\text{ID}}(\theta)} \leq \frac{\lambda_{\max}(H_{\text{OOD}})}{\lambda_{\min}(H_{\text{ID}})}.
\label{eq:layleigh_quotient}
\tag{11}
\end{equation}
where $\lambda_{\min}(H_{\text{ID}})>0$.

\bigskip
\subsection{Using Logistic Regression Hessian as a Covariance Approximation}\label{subsec:a_2}

\medskip

In logistic regression, the negative log-likelihood is given by
\[
-\ell(\theta) = -\sum_{i=1}^{n} \Bigl[ y_i \log \sigma(\theta^\top x_i) + (1-y_i)\log (1-\sigma(\theta^\top x_i)) \Bigr].
\]
Thus, the Hessian of the NCE (a variant of logistic regression) is upper bounded by the covariance matrix $\Sigma \in \mathbb{R}^{d \times d}$
\begin{equation}
\begin{split}
H &= \sum_{i=1}^{n} \sigma(\theta_0^\top x_i) \bigl(1-\sigma(\theta_0^\top x_i)\bigr) x_i x_i^\top \\
&\leq \frac{1}{4} \sum_{i=1}^{n} x_i x_i^\top = \frac{n}{4} \Sigma,
\end{split}
\tag{12}
\end{equation}
since \(\sigma(\theta_0^\top x_i)(1-\sigma(\theta_0^\top x_i)) \leq \frac{1}{4}\).

\medskip

\noindent Thus, it follows that
\begin{equation}
\frac{\text{NCE}_{\text{OOD}}}{\text{NCE}_{\text{ID}}} \leq \frac{\lambda_{\max}(H_{\text{OOD}})}{\lambda_{\min}(H_{\text{ID}})}
\leq \frac{\lambda_{\max}(\Sigma_{\text{OOD}})}{\lambda_{\min}(\Sigma_{\text{ID}})},
\tag{13}
\end{equation}
where $\Sigma_\textnormal{ID}$ and $\Sigma_\textnormal{OOD}$ are covariance matrices of data from ID and OOD, respectively.
\medskip

\noindent With the eigendecomposition \(\Sigma = Q\Lambda Q^\top\), the Frobenius norm is given by
\begin{equation}
\|\Sigma\|_F^2 = \operatorname{tr}(\Sigma \Sigma^\top) = \operatorname{tr}(\Lambda\Lambda^\top) = \sum (\text{eigenvalues})^2.
\tag{14}
\end{equation}
In addition, we have the inequality
\begin{equation}
\sqrt{\lambda_{\max}^2(\Sigma)} \leq \|\Sigma\|_F \leq \sqrt{\operatorname{rank}(\Sigma) \cdot \lambda^2_{\max}(\Sigma)}.
\label{eq:frobenius_norm_ineq}
\tag{15}
\end{equation}

\medskip

We assume that the ID and OOD covariance matrices satisfy the matching marginal condition, which means that the two distributions have identical marginal variances. In other words, the diagonal elements of their covariance matrices are the same, although the off-diagonal entries may differ. This condition preserves the variances of individual features across domains while allowing feature correlations to vary. In our study, this assumption is reasonable because we use normalized embeddings, which naturally align marginal variances. We empirically validate this condition on real datasets in Appendix~\ref{subsec:c_1}.
This condition is formalized as:
\[
\Sigma_{\text{OOD}} = \Sigma_{\text{ID}} + E,
\]
where $E \in \mathbb{R}^{d \times d}$ is a matrix that captures domain-specific differences. By assumption, E has zero diagonal entries and non-zero off-diagonal entries, meaning it only affects feature correlations while preserving individual feature variances.
By the triangle inequality,
\[
\|\Sigma_{\text{OOD}}\|_F \leq \|\Sigma_{\text{ID}}\|_F + \|E\|_F,
\]
and if \(\|E\|_F \leq \sqrt{d^2-d}\) (with \(|E_{ij}|\leq 1\) for \(i \neq j\)), one can bound the maximum singular value of \(\Sigma_{\text{OOD}}\). This leads to the bound with $\mathcal{L}_\text{OOD}$ and $\mathcal{L}_\textnormal{ID}$, which are losses of $\theta$ on OOD data and ID data, respectively.
\begin{equation}
\mathcal{L}_{\text{OOD}} \leq \mathcal{L}_{\text{ID}} + \frac{\lambda_{\max}(\Sigma_{\text{ID}})\times\sqrt{d} + \sqrt{d^2-d}}{\lambda_{\min}(\Sigma_{\text{ID}})}.
\label{eq:bound_of_Loss_OOD_appendix}
\tag{16}
\end{equation}

\bigskip
\subsection{Approximating Marginal Contributions of the Eigenvalue Term}\label{subsec:a_3}
\textbf{Problem Statement:} How can we use perturbation to compute the marginal value of a data point?

\medskip

\noindent Given a normalized embedding dataset \(\{x_1,x_2,\dots,x_n\}\) of ID, the covariance matrix is defined as
\begin{equation}
\Sigma_\textnormal{ID} = \frac{1}{n} \sum_{i=1}^{n} x_i x_i^\top.
\tag{17}
\end{equation}
When one data point \(x_k\) is removed, the new covariance matrix becomes
\begin{equation}
\Sigma_{-k} = \frac{1}{n-1} \sum_{i \neq k} x_i x_i^\top = \frac{n}{n-1}(\Sigma_\textnormal{ID} + \Delta_k)\approx\Sigma_\textnormal{ID} + \Delta_k,
\tag{18}
\end{equation}
where $\Delta_k = -\frac{1}{n}\, x_k x_k^\top, \frac{n}{n-1}\approx 1$.

\medskip

Let \(\lambda_{\max}(\Sigma_\textnormal{ID})\) and \(\lambda_{\min}(\Sigma_\textnormal{ID})\) be the maximum and minimum eigenvalues of \(\Sigma_\textnormal{ID}\), with corresponding normalized eigenvectors \(u_{\max}\) and \(u_{\min}\). A first-order perturbation yields
\[
\lambda_{\max}(\Sigma_{-k}) \approx \lambda_{\max}(\Sigma_\textnormal{ID}+\Delta_k) \approx \lambda_{\max}(\Sigma_\textnormal{ID}) + u_{\max}^\top \Delta_k\, u_{\max},
\]
\[
\lambda_{\min}(\Sigma_{-k}) \approx \lambda_{\min}(\Sigma_\textnormal{ID}+\Delta_k) \approx \lambda_{\min}(\Sigma_\textnormal{ID}) + u_{\min}^\top \Delta_k\, u_{\min}.
\]
Define
\[
\delta_{\max}^{(k)} := u_{\max}^\top \Delta_k\, u_{\max}, \quad \delta_{\min}^{(k)} := u_{\min}^\top \Delta_k\, u_{\min}.
\]

\medskip

\noindent Let \(f(\Sigma_\textnormal{ID})\) denote the approximated domain discrepancy function from Eq.~\ref{eq:bound_of_Loss_OOD_appendix}:
\[
f(\Sigma_\textnormal{ID}) = \frac{\lambda_{\max}(\Sigma_\textnormal{ID})\times\sqrt{d} + \sqrt{d^2-d}}{\lambda_{\min}(\Sigma_\textnormal{ID})}.
\]
After removing \(x_k\), we have
\begin{equation}
f(\Sigma_{-k}) \approx \frac{[\lambda_{\max}(\Sigma_\textnormal{ID})+\delta_{\max}^{(k)}]\times\sqrt{d} + \sqrt{d^2-d}}{\lambda_{\min}(\Sigma_\textnormal{ID})+\delta_{\min}^{(k)}}.
\tag{19}
\end{equation}

\medskip

\noindent Define
\[
A = \lambda_{\max}(\Sigma_\textnormal{ID})\times\sqrt{d} + \sqrt{d^2-d}, \quad B = \lambda_{\min}(\Sigma_\textnormal{ID}).
\]
A first-order expansion of the denominator gives:
\[
\frac{1}{B+\delta_{\min}^{(k)}} \approx \frac{1}{B}\left(1 - \frac{\delta_{\min}^{(k)}}{B}\right).
\]
Thus,
\begin{equation}
\begin{split}
f(\Sigma_{-k}) &\approx \frac{A + \sqrt{d}\times\delta_{\max}^{(k)}}{B}\left(1 - \frac{\delta_{\min}^{(k)}}{B}\right)\\[1mm]
&\approx \frac{A}{B} + \frac{\sqrt{d}\times\delta_{\max}^{(k)}}{B} - \frac{A\times\delta_{\min}^{(k)}}{B^2}.
\end{split}
\tag{20}
\end{equation}
Therefore, the change in the function, which approximates the marginal OOD-robust data value of \(x_k\), is
\begin{equation}
\begin{split}
&f(\Sigma_{-k}) - f(\Sigma_\textnormal{ID}) \approx \frac{\sqrt{d}\times\delta_{\max}^{(k)}}{B} - \frac{A\times\delta_{\min}^{(k)}}{B^2}\\[5mm]
&=\frac{\sqrt{d} \times \delta_{\max}^{(k)}}{\lambda_{\min}(\Sigma_\textnormal{ID})} 
- \frac{(\lambda_{\max}(\Sigma_\textnormal{ID}) \times \sqrt{d} + \sqrt{d^2 - d} \,) \times \delta_{\min}^{(k)}}{\lambda_{\min}(\Sigma_\textnormal{ID})^2} \nonumber.
\end{split}
\tag{21}
\end{equation}
The proposed term quantifies the marginal data value with respect to domain discrepancy, rather than the ID loss. Accordingly, it can be integrated into the marginal values derived from existing ID-based data valuation methods. Under the assumption that the OOD loss can be approximated by the sum of the ID loss and domain discrepancy, this enables principled data valuation in OOD settings.

The derivations above demonstrate how domain discrepancy can be bounded by the eigenvalue ratio of the Hessians, how the OOD covariance matrix can be related to the ID covariance matrix, and how perturbation analysis yields an approximation of the marginal data value.
In addition, we examine the potential bias arising from dominant eigenvalues, as previously discussed in PCA-related analyses~\cite{samadi2018price}.
Our formulation indicates that robustness improves when the spectral distribution is balanced, with sufficiently large minimum eigenvalues and no excessively dominant maximum eigenvalues.
This observation aligns with prior findings, suggesting that over-reliance on dominant directions can lead to biased representations, while a more uniform eigenvalue spectrum contributes to improved stability and generalization.

\section{Additional Experiment Setting}

\subsection{Dataset}\label{subsec:b_1}

\noindent\textbf{CIFAR-10.}
A widely used image classification dataset consisting of natural images from ten classes. We use it as the source domain for training.

\noindent\textbf{CIFAR-10 C.}
A corrupted version of CIFAR-10 that introduces common distribution shifts through 15 corruption types, each with multiple severity levels. We use the 5 severity level. CIFAR-10  serves as the target domain for evaluating robustness under distribution shift.

\noindent\textbf{VLCS.}
A domain generalization benchmark composed of four visual domains: VOC2007, LabelMe, Caltech101, and SUN09. In each evaluation setting, one domain is held out as the target while the model is trained on the remaining three. The target domain is rotated across all four domains.

\noindent\textbf{Amazon Reviews.}
A sentiment classification dataset organized by product category, with each category treated as a separate domain. We convert the 5-point rating into three sentiment classes (negative: 1–2, neutral: 3, positive: 4–5) and perform 3-class classification. Models are trained on one or more source categories and evaluated on a disjoint target category to assess cross-domain generalization.

\noindent\textbf{ImageNet.}
A large-scale image classification dataset with 1,000 classes. For scalability experiments, we use a subset of the training split. Robustness is measured under domain shifts. For this benchmark, we performed the data valuation experiment using a subset of 30,000 samples.

\noindent\textbf{DomainNet.}
A large-scale benchmark for multi-domain learning, containing six stylistically distinct domains: clipart (C), infograph (I), painting (P), quickdraw (Q), real (R), and sketch (S). We evaluate generalization by holding out one domain as the target and training on the remaining five. For this benchmark, we performed the data valuation experiment using a subset of 2,000 samples for each domain.

Unlike prior OOD-aware valuation methods that require access to OOD data during estimation~\cite{das2024checkselect, yoon2020data}, EV operates solely on ID data and therefore remains valid even when no OOD samples are available. This makes EV inherently robust to unseen OOD scenarios, a setting that is common in real-world deployments. All experiments use the train split of each dataset. Performance is also evaluated on randomly sampled data from the train split of the target domain (i.e., the domain not used for training), using a fixed random seed of 42 except for the instability ranking experiment.

\subsection{Experiment setting}\label{subsec:b_2}

\textbf{Evaluation protocols.}
We use three procedures.
\begin{itemize}
\item \textbf{Point addition.} We sample 2,000 in distribution examples. We compute values with each method. We form an initial training set of 1,000 and retrain while adding the highest value samples from the remaining pool. We evaluate on a different target domain.
\item \textbf{Data removal.} We sample 1,000 of 2,000 in distribution points. We score them, remove the top 50 percent by value, and train on the rest. We evaluate on the target domain. A larger drop in accuracy indicates a better ability to identify low utility samples.
\item \textbf{Instability.} We assess sensitivity to small changes in the training set. We fix 180 of 200 indices, resample the remaining 20, repeat valuation five times, and compute the standard deviation of value rankings on the fixed indices. These five runs use different random seeds (8, 18, 29, 39, 58).
\end{itemize}
\textbf{Baselines and parameters.}
In this work, we limit the EV-integrated methods to LAVA, KNN Shapley, and Data-OOB, selected based on a balance of performance and computational efficiency.
For KNN Shapley, we use a validation set of 1,000 examples and set the neighborhood size to 1,000. For Data-OOB, we follow the original paper with num models = 800. We train a logistic regression classifier for 10 epochs with a learning rate of 0.001.

\noindent\textbf{Hardware setting.}
We set seed 42 on a single RTX 4090 GPU and an Intel Xeon Gold 6426Y CPU with 32 cores.

\subsection{Weight parameter}
EV is combined with a baseline valuation score ($V_{\text{EV}}$). Since the EV term may have a different scale from other methods($V_{\text{base}}$), we center and scale it using the baseline statistics to make the two terms comparable. Specifically,
$\tilde V_{\text{EV}}
=\frac{V_{\text{EV}}-\mu_{\text{base}}}{\sigma_{\text{base}}},$
where $\mu_{\text{base}}=\mathrm{mean}(V_{\text{base}})$ and $\sigma_{\text{base}}=\mathrm{std}(V_{\text{base}})$.
The final score is
$V_{\text{final}} \;=\; V_{\text{base}} \;+\; w\,\tilde V_{\text{EV}}.$
In our experiments, we set w$\leq$1.

\section{Supplementary Experiments}
\subsection{Experiment on ImageNet and DomainNet}\label{subsec:c_1}
\setcounter{table}{2}

\begin{table*}[htbp]
\centering
\setlength{\tabcolsep}{1mm}
\begin{tabular}{l|cccc|cccccc}
\toprule
\textbf{Acc (\%)} ($\downarrow$)
& \multicolumn{4}{c|}{\textbf{ImageNet}} 
& \multicolumn{6}{c}{\textbf{DomainNet}} 
\\
\cmidrule(lr){1-1}\cmidrule(lr){2-5} \cmidrule(lr){6-11}
\textbf{Method} 
& V2 & S & R & A 
& C & I & P & Q & R & S 
\\
\midrule
Random               & 65.5        & 28.2        & 29.9        & 9.0        & 26.4   & 13.7  & 31.7  & 2.2   &43.8   & 18.6   \\
InfluenceFunction            & 65.6        & 28.3        & 29.5        & 8.7        & 25.6  & 12.6  & 30.9  & 2.2   & 41.2  & 19.7   \\
KNN Shapley                 & 40.3        & 18.0        & 17.0        & 7.8        & 21.5  & 9.5   & 24.7  & 2.7   & 34.0  & 14.2    \\
Data-OOB                  & 59.2        & 23.8        & 25.1        & 6.5        & 15.2  & 6.0   & 16.5  & 2.1   & 20.4  & 10.4       \\
\midrule
EV + KNN Shapley     &\textbf{40.3}&\textbf{17.9}&\textbf{16.9}& 7.8        & 20.2  & 9.2   & 23.7  & 2.6   & 33.0  & 13.5  \\
EV + Data-OOB       & 54.7        & 21.8        & 22.8        &\textbf{5.4}& \textbf{14.5}  & \textbf{5.8}   & \textbf{15.1}  & \textbf{2.0}  & \textbf{18.6}   & \textbf{10.2}  \\
\bottomrule
\end{tabular}
\caption{Data removal experiment. Train the model with 50\% of the data, which is the lowest data value in the ID set, and evaluate performance on different domain data. \textbf{Lower is better.} Across both large and real benchmarks, EV augmented variants consistently achieve the lowest error, which means EV achieves stronger OOD robustness than other methods. Because of their prohibitive time complexity on large, high-cardinality datasets, LAVA and Deviation are omitted. }
\label{tab:data_removal_real_appendix}
\end{table*}
\noindent We extend the Data Removal experiment from the main paper to more challenging benchmarks. For ImageNet, data valuation was conducted on 30,000 training samples from the train split of ImageNet. For DomainNet, 2,000 samples were drawn from each domain, and data valuation was performed using the remaining 10,000 samples, excluding the target domain. The experimental setup follows that of Table 2 in the main paper.
As shown in Table~\ref{tab:data_removal_real_appendix}, EV continues to outperform other methods in OOD domains, and the performance gain from integrating EV is consistently observed compared to the base methods without EV. While Table 2 reports the averaged performance over DomainNet due to space constraints, per-domain results in the appendix also confirm that EV consistently improves performance across individual domains.
 Notably, on V2, EV + KNN Shapley slightly outperforms KNN Shapley alone, even at the second decimal place. Due to computational constraints, LAVA was excluded due to its sensitivity to the number of labels, and Deviation was excluded because it scales poorly with dataset size.

\subsection{Experiment under Indefinite Hessian}\label{subsec:c_2}
When the Hessian is indefinite, Eq.~\ref{eq:layleigh_quotient} does not directly apply because the minimum eigenvalue can be negative.
By following the same derivation as in the positive semi-definite (PSD) case and restricting the analysis to the subspace spanned by negative eigenvectors, a Rayleigh–based upper bound can still be obtained, where the bound depends on the largest negative eigenvalue.
For simplicity, we use the same formulation as in the PSD case.
Empirically, incorporating EV consistently improved performance even when the Hessian was indefinite.
Table~\ref{tab:cifar10c_MLP} reports the results under the same setup as Table 1 using a 2-layer MLP.
\begin{table}[t]
\centering
\begin{tabular}{l|c}
\toprule
\textbf{Method} & \textbf{CIFAR-10 C} $\downarrow$ \\
\midrule
Random      & 51.9 \\
InfluenceFunction   & 50.7 \\
Deviation   & 51.0 \\
LAVA        & 44.5 \\
KNN Shapley        & 41.3 \\
Data-OOB         & 51.8 \\
\midrule
EV + LAVA   & \textbf{40.1} \\
EV + KNN Shapley    & 40.2 \\
EV + Data-OOB    & 44.0 \\
\bottomrule
\end{tabular}
\caption{Data removal experiment. Comparison of data valuation methods on CIFAR-10 C using a 2-layer MLP model under domain shift. Lower accuracy indicates more effective identification of low-utility samples. The experimental setup, except the model, follows the same configuration as described in Table 1.}
\label{tab:cifar10c_MLP}
\end{table}

\subsection{Experiment under Different Embedder}\label{subsec:c_3}
Our experimental pipeline relies on embedding representations, which raises a natural concern that the performance of EV might depend on the quality or choice of the embedder. To examine this, we repeated the CIFAR-10 data-removal experiments (Table 1) using a variety of embedding models. As reported in Table~\ref{tab:data_removal_different_embedder}, EV consistently maintains strong robustness to OOD data across different embedders. This result indicates that the OOD robustness of EV is not tied to a specific embedding model and that EV remains effective even when alternative feature extractors are used.

\begin{table}[t]
\centering
\resizebox{0.49\textwidth}{!}{%
\begin{tabular}{@{}l|c|c|c|c|c@{}}  
\toprule
Method / Embedder & ResNet-18 & ResNet-50 & ViT-B/16 & ViT-L/16 & \textbf{Avg} \\
\midrule
Random          & 48.7 & 47.0 & 67.0 & 70.9 & 58.7 \\
InfluenceFunction       & 46.3 & 47.8 & 67.5 & 70.2 & 57.9 \\
Deviation       & 43.6 & 46.5 & 66.3 & 71.6 & 57.0 \\
LAVA            & 42.7 & 46.3 & 66.6 & 69.9 & 56.4 \\
KNN Shapley            & 33.8 & 38.8 & 60.4 & 63.3 & 49.1 \\
Data-OOB             & 36.3 & 44.6 & 67.8 & 70.4 & 54.8 \\
\midrule
EV + LAVA & 42.4 & 43.9 & \textbf{58.1} & \textbf{58.6} & 50.8 \\
EV + KNN Shapley & 33.9 & \textbf{38.6} & 60.3 & 62.5 & \textbf{48.8} \\
EV + Data-OOB  & \textbf{33.6} & 43.6 & 63.9 & 66.6 & 51.9 \\

\bottomrule
\end{tabular}
}
\caption{Data removal experiment. Comparison of data valuation methods on CIFAR-10 C using different embedders. Except for the choice of embedder, the experimental setup follows the same configuration as in Table 1.}
\label{tab:data_removal_different_embedder}
\end{table}
\subsection{Different Removal Rate Results on CIFAR-10 C}\label{subsec:c_5}
While Table 1 reports the results of removing the top 50\% highest-value samples, we additionally evaluate a less extreme setting to show that EV is not effective only under large removal rates. Table~\ref{tab:cifar10c_10percent} presents the results when the removal rate is reduced to 10\%. EV continues to deliver strong performance at this lower rate, exhibiting consistent improvements across methods. These results indicate that EV is broadly applicable and remains effective under a range of removal rates.

\begin{table}[t]
\centering
\begin{tabular}{l|c}
\toprule
\textbf{Method} & \textbf{CIFAR-10 C} $\downarrow$ \\
\midrule
Random      & 50.1 \\
InfluenceFunction   & 49.3 \\
Deviation   & 50.7 \\
LAVA        & 50.03 \\
KNN Shapley        & 48.19 \\
Data-OOB         & 48.5 \\
\midrule
EV + LAVA   & 48.6 \\
EV + KNN Shapley    & \textbf{45.8} \\
EV + Data-OOB    & 46.9 \\
\bottomrule
\end{tabular}
\caption{Data removal experiment. Comparison of data valuation methods on CIFAR-10 C using different removal rates (10\%). Lower accuracy indicates more effective identification of low-utility samples. The experimental setup, except removal rate, follows the same configuration as described in Table 1.}
\label{tab:cifar10c_10percent}
\end{table}

\subsection{Additional Baseline Results on CIFAR-10 C}\label{subsec:c_4}
We extended the data removal experiment of Table 1 by applying the same evaluation procedure to additional baselines on CIFAR-10 C.
As summarized in Table~\ref{tab:cifar10c_additional_baseline}, we included both DAVINZ~\cite{wu2022davinz} and Banzhaf~\cite{wang2023data} under the identical experimental setup.
Across these baselines, incorporating EV led to consistent improvements, indicating that the proposed approach effectively enhances the ability of diverse valuation methods to identify low-utility samples.
These results confirm that the benefit of EV is not limited to a specific algorithm but generalizes across methods with different underlying principles.

\begin{table}[t]
\centering
\begin{tabular}{l|c}
\toprule
\textbf{Method} & \textbf{CIFAR-10 C} $\downarrow$ \\
\midrule
Random      & 47.0 \\
InfluenceFunction   & 47.8 \\
Deviation   & 46.5 \\
LAVA        & 46.3 \\
KNN Shapley        & 38.8 \\
Data-OOB         & 44.6 \\
Banzhaf & 46.7\\
DAVINZ & 47.3\\
\midrule
EV + LAVA   & 43.9 \\
EV + KNN Shapley    & 38.6 \\
EV + Data-OOB    & 43.6 \\
EV + Banzhaf   & 45.4 \\
EV + DAVINZ    & \textbf{35.6} \\
\bottomrule
\end{tabular}
\caption{Data removal experiment. Comparison of data valuation methods on CIFAR-10-C under domain shift, including additional results for Banzhaf and DAVINZ methods. The experimental setup follows the same configuration as described in Table 1.}
\label{tab:cifar10c_additional_baseline}
\end{table}

\subsection{Additional metrics for the Instability Ranking experiment}\label{subsec:c_6}

In Figure 4, we examine how sensitive each valuation method is to small perturbations in the training subset by computing the standard deviation of the resulting data value rankings across multiple resamples. This analysis illustrates how much the ranking produced by each method fluctuates when the underlying subset used for valuation is slightly altered. Table~\ref{tab:instability_ranking_additional_metric} provides a more comprehensive evaluation based on the rankings obtained in Figure 4. In addition to reporting the standard deviation, we also compute the variance, Spearman correlation, and Kendall correlation across repeated runs. These complementary metrics reveal that EV achieves substantially greater stability than Deviation, demonstrating that EV retains robust OOD performance even when the training subset undergoes minor changes.

\begin{table}[t]
\centering
\resizebox{0.49\textwidth}{!}{%
\begin{tabular}{@{}l|c|c|c|c}  
\toprule
Method / Metric & Std. $(\downarrow)$& Variance $(\downarrow)$ & Spearman $(\uparrow)$ & Kendall $(\uparrow)$ \\
\midrule
Random          & 48.58 & 2582.65 & 0.05 & 0.03  \\
Deviation       & 34.25 & 1489.68 & 0.45 & 0.32 \\
\midrule
EV + LAVA       & \textbf{7.60} & \textbf{75.45} & \textbf{0.97} & \textbf{0.86} \\
EV + KNN Shapley  & 7.84 & 81.77 & \textbf{0.97} & \textbf{0.86} \\
EV + Data-OOB   & 19.18 & 592.09 & 0.77 & 0.62 \\

\bottomrule
\end{tabular}
}
\caption{Additional metrics for the Instability Ranking experiment}
\label{tab:instability_ranking_additional_metric}
\end{table}

\end{document}